\documentclass{article}

\usepackage{times}
\usepackage{graphicx} % more modern
\usepackage{subfigure}

\usepackage{natbib}

\usepackage{algorithm}
\usepackage{algorithmic}

\usepackage{amsmath}
\usepackage{amsfonts}
\usepackage{mathtools}

\usepackage{flushend}

\usepackage{hyperref}

\usepackage[accepted]{icml2017}

\newcommand{\cmt}[1]{{\footnotesize\textcolor{red}{#1}}}

\newcommand{\todo}[1]{\cmt{TO-DO: #1}}

\long\def\ignorethis#1{}

\newcommand{\reward}{R}

\newcommand{\action}{\mathbf{a}}

\icmltitlerunning{Model-Agnostic Meta-Learning for Fast Adaptation of Deep Networks}

\begin{document}

\twocolumn[
\icmltitle{Model-Agnostic Meta-Learning for Fast Adaptation of Deep Networks}

% It is OKAY to include author information, even for blind
% submissions: the style file will automatically remove it for you
% unless you've provided the [accepted] option to the icml2016
% package.
\icmlsetsymbol{equal}{*}

\begin{icmlauthorlist}
\icmlauthor{Chelsea Finn}{be}
\icmlauthor{Pieter Abbeel}{be,openai}
\icmlauthor{Sergey Levine}{be}
\end{icmlauthorlist}

\icmlaffiliation{be}{University of California, Berkeley}
\icmlaffiliation{openai}{OpenAI}
%\icmlaffiliation{ed}{University of Edenborrow, Edenborrow, United Kingdom}

\icmlcorrespondingauthor{Chelsea Finn}{cbfinn@eecs.berkeley.edu}

% You may provide any keywords that you
% find helpful for describing your paper; these are used to populate
% the "keywords" metadata in the PDF but will not be shown in the document
\icmlkeywords{meta-learning, deep learning}

\vskip 0.3in
]

\printAffiliationsAndNotice{}

\begin{abstract}
We propose an algorithm for meta-learning that is model-agnostic, in the sense that it is compatible with any model trained with gradient descent and applicable to a variety of different learning problems, including classification, regression, and reinforcement learning. The goal of meta-learning is to train a model on a variety of learning tasks, such that it can solve new learning tasks using only a small number of training samples. In our approach, the parameters of the model are explicitly trained such that a small number of gradient steps with a small amount of training data from a new task will produce good generalization performance on that task. In effect, our method trains the model to be easy to fine-tune. We demonstrate that this approach leads to state-of-the-art performance on two few-shot image classification benchmarks, produces good results on few-shot regression, and accelerates fine-tuning for policy gradient reinforcement learning with neural network policies.
\end{abstract}

\section{Introduction}
\label{sec:intro}

% This paragraph is for motivation.
Learning quickly is a hallmark of human intelligence, whether it involves recognizing objects from a few examples or quickly learning new skills
after just minutes of experience. Our artificial agents should be able to do the same, learning and adapting quickly from only a few examples, and continuing to adapt as more data becomes available. This kind of fast and flexible learning is challenging, since the agent must integrate its prior experience with a small amount of new information, while avoiding overfitting to the new data. Furthermore, the form of prior experience and new data will depend on the task. As such, for the greatest applicability, the mechanism for learning to learn (or meta-learning) should be general to the task and the form of computation required to complete the task.

% This paragraph is for describing our method and what differentiates it from prior methods.
In this work, we propose a meta-learning algorithm that is general and model-agnostic, in the sense that it can be directly applied to any learning problem and model that is trained with a gradient descent procedure. Our focus is on deep neural network models, but we illustrate how our approach can easily handle different architectures and different problem settings, including classification, regression, and policy gradient reinforcement learning, with minimal modification.
In meta-learning, the goal of the trained model is to quickly learn a new task from a small amount of new data, and the model is
trained by the meta-learner to be able to learn on a large number of different tasks.
The key idea underlying our method is to
train the model's initial parameters such that the model has maximal performance on a new task after the parameters have been updated
through one or more gradient steps computed with a small amount of data from that new task.
Unlike prior meta-learning methods that learn an update function or learning rule~\cite{schmidhuber1987,bengiobengio1,learntolearnbygdbygd,hugo}, our algorithm does not expand the number of learned parameters nor place constraints on the model architecture (e.g. by requiring a recurrent model~\cite{mann} or a Siamese network~\cite{siameseoneshot}), and it can be readily combined with fully connected, convolutional, or recurrent neural networks. It can also be used with a variety of loss functions, including differentiable supervised losses and non-differentiable reinforcement learning objectives. % would be good to cite matching networks somewhere.

%%SL.02.18: added this paragraph
The process of training a model's parameters such that a few gradient steps, or even a single gradient step, can produce good results on a new task can be viewed from a feature learning standpoint as building an internal representation that is broadly suitable for many tasks. If the internal representation is suitable to many tasks, simply fine-tuning the parameters slightly (e.g. by primarily modifying the top layer weights in a feedforward model) can produce good results. In effect, our procedure optimizes for models that are easy and fast to fine-tune, allowing the adaptation to happen in the right space for fast learning. From a dynamical systems standpoint, our learning process can be viewed as maximizing the sensitivity of the loss functions of new tasks with respect to the parameters: when the sensitivity is high, small local changes to the parameters can lead to large improvements in the task loss.

The primary contribution of this work is a simple model- and task-agnostic algorithm for meta-learning that trains a model's parameters such that a small number of gradient updates will lead to fast learning on a new task.
We demonstrate the algorithm on different model types, including fully connected and convolutional networks, and in several distinct domains, including few-shot regression, image classification, and reinforcement learning.
Our evaluation shows that our meta-learning algorithm compares favorably to state-of-the-art one-shot learning methods designed specifically for supervised classification, while using fewer parameters, but that it can also be readily applied to regression and can accelerate reinforcement learning in the presence of task variability, substantially outperforming direct pretraining as initialization.

\section{Model-Agnostic Meta-Learning}
\label{sec:overview}

We aim to train models that can achieve rapid adaptation, a problem setting that is often formalized as few-shot learning. In this section, we will define the problem setup and present the general form of our algorithm.

\subsection{Meta-Learning Problem Set-Up}
\label{sec:problem}

\newcommand{\task}{\mathcal{T}}
\newcommand{\loss}{\mathcal{L}}
\newcommand{\inp}{\mathbf{x}}
\newcommand{\learner}{f}
\newcommand{\lossi}{\loss_{\task_i}}

The goal of few-shot meta-learning
is to train a model that can quickly adapt to a new task using only a few datapoints and training iterations.
To accomplish this, the model or learner is trained during a meta-learning phase on a set of tasks, such that the trained model can quickly adapt to new tasks using only a small number of examples or trials.
In effect, the meta-learning problem treats entire tasks as training examples.
In this section, we formalize this meta-learning problem setting in a general manner, including brief examples of different learning domains.
We will discuss two different learning domains in detail in Section~\ref{sec:instantiations}.

We consider a model, denoted $\learner$, that maps observations $\inp$ to outputs $\action$.
During meta-learning, the model
is trained to be able to adapt to a large or infinite number of tasks.
Since we would like to apply our framework to a variety of learning problems, from classification to reinforcement learning, we introduce a generic notion of a learning task below.
Formally, each task $\task = \{ \loss(\inp_1,\action_1,\dots,\inp_H,\action_H), q(\inp_1), q(\inp_{t+1} | \inp_t, \action_t), H  \}$
consists of a loss function $\loss$, a distribution over initial observations $q(\inp_1)$, a transition distribution $ q(\inp_{t+1} | \inp_t, \action_t)$, and an episode length $H$. In i.i.d. supervised learning problems, the length $H\!=\!1$.
The model may generate samples of length $H$ by choosing an output $\action_t$ at each time $t$. 
%Task-specific feedback is provided by 
The loss $\loss(\inp_1,\action_1,\dots,\inp_H,\action_H)\rightarrow \mathbb{R}$, provides task-specific feedback, which might be in the form of a misclassification loss or a cost function in a Markov decision process.

In our meta-learning scenario, we consider a distribution over tasks $p(\task)$ that we want our model to be able to adapt to.
In the $K$-shot learning setting, the model is trained to learn a new task $\task_i$ drawn from $p(\task)$ from only $K$ samples drawn from $q_i$ and feedback $\lossi$ generated by $\task_i$.
During meta-training, a task $\task_i$ is sampled from $p(\task)$, the model is trained with $K$ samples and feedback from the corresponding loss $\lossi$ from $\task_i$, and then tested on new samples from $\task_i$.
The model $\learner$ is then improved by considering how the \emph{test} error on new data from $q_i$ changes with respect to the parameters. In effect, the test error on sampled tasks $\task_i$ serves as the training error of the meta-learning process.
%After one task or a batch of tasks, the meta-parameters are updated to maximize the model's test performance over the tasks.
At the end of meta-training, new tasks are sampled from $p(\task)$,
and meta-performance is measured by the model's performance after learning from $K$ samples.
Generally, tasks used for meta-testing are held out during meta-training.

\subsection{A Model-Agnostic Meta-Learning Algorithm}
\label{sec:maml}
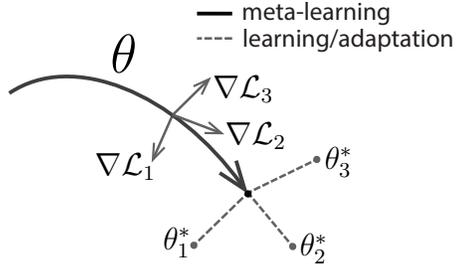
\begin{figure}
\setlength{\unitlength}{0.5\columnwidth}
\begin{picture}(1.99,0.9) \linethickness{0.5pt}
\put(0.25,-0.1){\includegraphics[width=0.75\columnwidth]{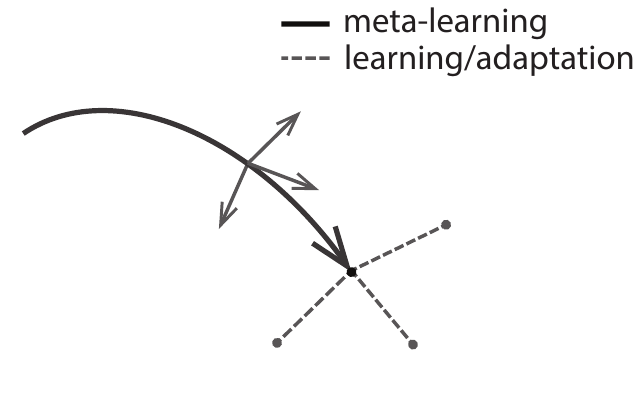}}
\put(0.63,0.63){\huge$\theta$}
\put(0.58,0.3){\large$\nabla \mathcal{L}_1$}
\put(1.0,0.4){\large$\nabla \mathcal{L}_2$}
\put(0.96,0.55){\large$\nabla \mathcal{L}_3$}

\put(0.80,0.06){\large$\theta^*_1$}
\put(1.24,0.04){\large$\theta^*_2$}
\put(1.32,0.32){\large$\theta^*_3$}
\end{picture}
\vspace{-0.8cm}
\caption{
Diagram of our model-agnostic meta-learning algorithm (MAML), which optimizes for a representation $\theta$ that can quickly adapt to new tasks.
\label{fig:teaser}
\vspace{-0.5cm}
}
\end{figure}

In contrast to prior work, which has sought to train recurrent neural networks that ingest entire datasets~\cite{mann,rl2} or feature embeddings that can be combined with nonparametric methods at test time~\cite{matchingnets,siameseoneshot}, we propose a method that can learn the parameters of any standard model via meta-learning in such a way as to prepare that model for fast adaptation. The intuition behind this approach is that some internal representations are more transferrable than others. For example, a neural network might learn internal features that are broadly applicable to all tasks in $p(\task)$, rather than a single individual task. How can we encourage the emergence of such general-purpose representations? We take an explicit approach to this problem: since the model will be fine-tuned using a gradient-based learning rule on a new task, we will aim to learn a model in such a way that this gradient-based learning rule can make rapid progress on new tasks drawn from $p(\task)$, without overfitting. In effect, we will aim to find model parameters that are \emph{sensitive} to changes in the task, such that small changes in the parameters will produce large improvements on the loss function of any task drawn from $p(\task)$, when altered in the direction of the gradient of that loss (see Figure~\ref{fig:teaser}). We make no assumption on the form of the model, other than to assume that it is parametrized by some parameter vector $\theta$, and that the loss function is smooth enough in $\theta$ that we can use gradient-based learning techniques.

Formally, we consider a model
represented by a parametrized function $\learner_\theta$
with parameters $\theta$.
When adapting to a new task $\task_i$, the model's parameters $\theta$ become $\theta_i'$.
% maybe note here that the new parameters are task specific.
In our method, the updated parameter vector $\theta_i'$ is computed using one or more gradient descent updates on task $\task_i$.
For example, when using one gradient update,
\vspace{-0.15cm}
$$
\vspace{-0.15cm}
\theta_i'=\theta-\alpha \nabla_\theta  \lossi(  \learner_\theta ).
$$
The step size $\alpha$ may be fixed as a hyperparameter or meta-learned.
For simplicity of notation, we will consider one gradient update for the rest of this section, but using multiple gradient updates is a straightforward extension.

The model parameters are trained by optimizing for the performance of $\learner_{\theta_i'}$ with respect to $\theta$ across tasks sampled from $p(\task)$.
More concretely, the meta-objective is as follows:
\vspace{-0.15cm}
\begin{align*}
\vspace{-0.2cm}
\min_\theta \sum_{\task_i \sim p(\task)}  \lossi ( \learner_{\theta_i'}) 
= \sum_{\task_i \sim p(\task)}  \lossi ( \learner_{\theta - \alpha \nabla_\theta \lossi(f_\theta)})
\end{align*}
% \sum_{\inp_{1:H} \sim \task_i} \inp_{1:H},
Note that the meta-optimization is performed over the model parameters $\theta$, whereas the objective is computed using the updated model parameters $\theta'$.
In effect, our proposed method aims to optimize the model parameters such that one or a small number of gradient steps on a new task will produce maximally effective behavior on that task.

The meta-optimization across tasks is performed via stochastic gradient descent (SGD), such that the model parameters $\theta$ are updated as follows:
\vspace{-0.15cm}
\begin{equation}
\label{eq:metaupdate}
\vspace{-0.2cm}
\theta \leftarrow \theta - \beta \nabla_\theta \sum_{\task_i \sim p(\task)}  \lossi ( \learner_{\theta_i'})
\end{equation}
where $\beta$ is the meta step size. The full algorithm, in the general case, is outlined in Algorithm~\ref{alg:maml}.

\begin{algorithm}[t]
\caption{Model-Agnostic Meta-Learning}
\label{alg:maml}
\begin{algorithmic}[1]
\REQUIRE $p(\task)$: distribution over tasks
\REQUIRE $\alpha$, $\beta$: step size hyperparameters
\STATE randomly initialize $\theta$
\WHILE{not done}
\STATE Sample batch of tasks $\task_i \sim p(\task)$
  \FORALL{$\task_i$}
 \STATE Evaluate $\nabla_\theta \lossi(\learner_\theta)$ with respect to $K$ examples
 \STATE Compute adapted parameters with gradient descent: $\theta_i'=\theta-\alpha \nabla_\theta  \lossi(  \learner_\theta )$
 \ENDFOR
 \STATE Update $\theta \leftarrow \theta - \beta \nabla_\theta \sum_{\task_i \sim p(\task)}  \lossi ( \learner_{\theta_i'})$
\ENDWHILE
%\STATE while 
\end{algorithmic}
\end{algorithm}

The MAML meta-gradient update involves a gradient through a gradient. Computationally, this requires an additional backward pass through $f$ to compute Hessian-vector products, which is supported by standard deep learning libraries such as TensorFlow~\cite{tensorflow}. In our experiments, we also include a comparison to dropping this backward pass and using a first-order approximation, which we discuss in Section~\ref{sec:image_results}.

\vspace{-0.15cm}
\section{Species of MAML}
%\section{Instantiations of MAML}
\label{sec:instantiations}
\vspace{-0.15cm}

In this section, we discuss specific instantiations of our meta-learning algorithm for supervised learning and reinforcement learning. The domains differ in the form of loss function and in how data is generated by the task and presented to the model, but the same basic adaptation mechanism can be applied in both cases.

\vspace{-0.1cm}
\subsection{Supervised Regression and Classification}

\newcommand{\target}{\mathbf{y}}

% discuss high level applications of few-shot supervised tasks
Few-shot learning is well-studied in the domain of supervised tasks, where the goal is to learn a new function from only a few input/output pairs for that task, using prior data from similar tasks for meta-learning. For example, the goal might be to classify images of a Segway after seeing only one or a few examples of a Segway, with a model that has previously seen many other types of objects. Likewise, in few-shot regression, the goal is to predict the outputs of a continuous-valued function from only a few datapoints sampled from that function, after training on many functions with similar statistical properties.

To formalize the supervised regression and classification problems in the context of the meta-learning definitions in Section~\ref{sec:problem}, we can define the horizon
 $H=1$ and drop the timestep subscript on $\inp_t$, since the model accepts a single input and produces a single output, rather than a sequence of inputs and outputs. The task $\task_i$ generates $K$ i.i.d. observations $\inp$ from $q_i$, and the task loss is represented by the error between the model's output for $\inp$ and the corresponding target values $\target$ for that observation and task.
  
Two common loss functions used for supervised classification and regression are cross-entropy and mean-squared error (MSE), which we will describe below; though, other supervised loss functions may be used as well. For regression tasks using mean-squared error, the loss takes the form:
\vspace{-0.15cm}
\begin{align}
\vspace{-0.2cm}
\label{eq:sup1}
\lossi(\learner_\phi ) = \!\!\!\!\!\! \sum_{\inp^{(j)}, \target^{(j)} \sim \task_i}  \lVert \learner_\phi(\inp^{(j)}) - \target^{(j)}  \rVert_2^2,
\end{align}
where $\inp^{(j)}, \target^{(j)}$ are an input/output pair sampled from task $\task_i$. In $K$-shot regression tasks, $K$ input/output pairs are provided for learning for each task.

Similarly, for discrete classification tasks with a cross-entropy loss, the loss takes the form:
% TODO - have another sum over classes?
\vspace{-0.15cm}
\begin{equation}
\begin{aligned}
\label{eq:sup2}
\lossi( \learner_\phi ) = \!\!\!\!\!\! \sum_{\inp^{(j)}, \target^{(j)} \sim \task_i} \!\!\!\!\!\! &\target^{(j)} \log \learner_\phi(\inp^{(j)}) \\
&+(1-\target^{(j)}) \log (1-\learner_\phi(\inp^{(j)}))
\end{aligned}
\end{equation}
According to the conventional terminology, $K$-shot classification tasks use $K$ input/output pairs from each class, for a total of $NK$ data points for $N$-way classification. Given a distribution over tasks $p(\task_i)$, these loss functions can be directly inserted into the equations in Section~\ref{sec:maml} to perform meta-learning, as detailed in Algorithm~\ref{alg:mamlsup}.

% This algorithm has maybe a few loose ends, with D and D'.
\begin{algorithm}[t]
\caption{MAML for Few-Shot Supervised Learning}
\label{alg:mamlsup}
\begin{algorithmic}[1]
{\footnotesize
\REQUIRE $p(\task)$: distribution over tasks
\REQUIRE $\alpha$, $\beta$: step size hyperparameters
\STATE randomly initialize $\theta$
\WHILE{not done}
\STATE Sample batch of tasks $\task_i \sim p(\task)$
  \FORALL{$\task_i$}
      \STATE Sample $K$ datapoints $\mathcal{D}=\{\inp^{(j)}, \target^{(j)}\}$ from $\task_i$
      \STATE Evaluate $\nabla_\theta \lossi(\learner_\theta)$ using $\mathcal{D}$ and $\lossi$ in Equation~(\ref{eq:sup1}) or~(\ref{eq:sup2})
      \STATE Compute adapted parameters with gradient descent: $\theta_i'=\theta-\alpha \nabla_\theta  \lossi(  \learner_\theta )$
      \STATE Sample datapoints $\mathcal{D}_i'=\{\inp^{(j)}, \target^{(j)}\}$ from $\task_i$ for the meta-update
 \ENDFOR
 \STATE Update $\theta \leftarrow \theta - \beta \nabla_\theta \sum_{\task_i \sim p(\task)}  \lossi ( \learner_{\theta_i'})$ using each $\mathcal{D}_i'$ and $\lossi$ in Equation~\ref{eq:sup1} or~\ref{eq:sup2}
\ENDWHILE
}
%\STATE while 
\end{algorithmic}
\end{algorithm}
%%SL.02.23: be careful with pseudocode placement, I was very confused b/c I was reading RL pseudocode in the SL section

\subsection{Reinforcement Learning}

In reinforcement learning (RL), the goal of few-shot meta-learning is to enable an agent to quickly acquire a policy for a new test task using only a small amount of experience in the test setting. A new task might involve achieving a new goal or succeeding on a previously trained goal in a new environment. For example, an agent might learn to quickly figure out how to navigate mazes so that, when faced with a new maze, it can determine how to reliably reach the exit with only a few samples.
In this section, we will discuss how MAML can be applied to meta-learning for RL.

Each RL task $\task_i$ contains an initial state distribution $q_i(\inp_1)$ and a transition distribution $q_i(\inp_{t+1}|\inp_t,\action_t)$, and the loss $\lossi$ corresponds to the (negative) reward function $\reward$. The entire task is therefore a Markov decision process (MDP) with horizon $H$, where the learner is allowed to query a limited number of sample trajectories for few-shot learning. Any aspect of the MDP may change across tasks in $p(\task)$. The model being learned, $\learner_\theta$, is a policy that maps from states $\inp_t$ to a distribution over actions $\action_t$ at each timestep $t \in \{1,...,H\}$. The loss for task $\task_i$ and model $\learner_\phi$ takes the form
\vspace{-0.1cm}
\begin{align}
\label{eq:rl}
\lossi( \learner_\phi) = - \mathbb{E}_{\inp_t, \action_t \sim \learner_\phi, q_{\task_i}}\left[ \sum_{t=1}^H \reward_i(\inp_t, \action_t)  \right].
\end{align}
In $K$-shot reinforcement learning, $K$ rollouts from $\learner_\theta$ and
task $\task_i$, $(\inp_1,\action_1,...\inp_H)$, and the corresponding rewards $\reward(\inp_t, \action_t)$, may be used for adaptation on a new task $\task_i$. Since the expected reward is generally not differentiable due to unknown dynamics, we use policy gradient methods to estimate the gradient both for the model gradient update(s) and the meta-optimization. Since policy gradients are an on-policy algorithm, each additional gradient step during the adaptation of $f_\theta$ requires new samples from the current policy $f_{\theta_{i'}}$. We detail the algorithm in Algorithm~\ref{alg:mamlrl}. This algorithm has the same structure as Algorithm~\ref{alg:mamlsup}, with the principal difference being that steps 5 and 8 require sampling trajectories from the environment corresponding to task $\task_i$. Practical implementations of this method may also use a variety of improvements recently proposed for policy gradient algorithms, including state or action-dependent baselines and trust regions~\cite{trpo}.

\begin{algorithm}[t]
\caption{MAML for Reinforcement Learning}
\label{alg:mamlrl}
\begin{algorithmic}[1]
{\footnotesize
\REQUIRE $p(\task)$: distribution over tasks
\REQUIRE $\alpha$, $\beta$: step size hyperparameters
\STATE randomly initialize $\theta$
\WHILE{not done}
\STATE Sample batch of tasks $\task_i \sim p(\task)$
  \FORALL{$\task_i$}
      \STATE Sample $K$ trajectories $\mathcal{D}=\{(\inp_1,\action_1,...\inp_H)\}$ using $f_\theta$ in $\task_i$
      \STATE Evaluate $\nabla_\theta \lossi(\learner_\theta)$ using $\mathcal{D}$ and $\loss_{\task_i}$ in Equation~\ref{eq:rl}
      \STATE Compute adapted parameters with gradient descent: $\theta_i'=\theta-\alpha \nabla_\theta  \loss_{\task_i}(  \learner_\theta )$
      \STATE Sample trajectories $\mathcal{D}_i'=\{(\inp_1,\action_1,...\inp_H)\}$ using $f_{\theta_i'}$ in $\task_i$ 
 \ENDFOR
 \STATE Update $\theta \leftarrow \theta - \beta \nabla_\theta \sum_{\task_i \sim p(\task)}  \loss_{\task_i} ( \learner_{\theta_i'})$ using each $\mathcal{D}_i'$ and $\loss_{\task_i}$ in Equation~\ref{eq:rl}
\ENDWHILE
}
%\STATE while 
\end{algorithmic}
\end{algorithm}

\section{Related Work}

The method that we propose in this paper addresses the general problem of meta-learning~\cite{thrun,schmidhuber1987,naik}, which includes few-shot learning.
A popular approach for meta-learning is to train a meta-learner that learns how to update the parameters of the learner's model~\cite{bengiobengio1,schmidfastweights,bengiobengio2}. This approach has been applied to learning to optimize deep networks~\cite{hochreiter,learntolearnbygdbygd,learntooptimize}, as well as for learning dynamically changing recurrent networks~\cite{hypernets}. 
%Similar methods have also been proposed that use evolutionary algorithms~\cite{evolvesearch}. 
One recent approach learns both the weight initialization and the optimizer, for few-shot image recognition~\cite{hugo}. Unlike these methods, the MAML learner's weights are updated using the gradient, rather than a learned update; our method does not introduce additional parameters for meta-learning nor require a particular learner architecture.

Few-shot learning methods have also been developed for specific tasks such as generative modeling~\cite{neuralstatistician,oneshotgenicml} and image recognition~\cite{matchingnets}. One successful approach for few-shot classification is to learn to compare new examples in a learned metric space using e.g. Siamese networks~\cite{siameseoneshot} or recurrence with attention mechanisms~\cite{matchingnets,comparators,prototypical}.
These approaches have generated some of the most successful results, but are difficult to directly extend to other problems, such as reinforcement learning. Our method, in contrast, is agnostic to the form of the model and to the particular learning task.

Another approach to meta-learning is to train memory-augmented models on many tasks, where 
%each episode corresponds to a new task and
the recurrent learner is trained to adapt to new tasks as it is rolled out. Such networks have been applied to few-shot image recognition~\cite{mann,metanets} and learning ``fast'' reinforcement learning agents~\cite{rl2,learningrl}.
%Memory augmentation can also take the form of a recurrent network with fast Hebbian learning updates~\cite{fastweights}. %This latter approach is inspired by the idea of using ``fast weights'' in addition to the standard ``slow'' weights in neural networks~\cite{hinton1987,malsburg}.
Our experiments show that our method outperforms the recurrent approach on few-shot classification. Furthermore, unlike these methods, our approach simply provides a good weight initialization and uses the same gradient descent update for both the learner and meta-update.
As a result, it is straightforward to finetune the learner for additional gradient steps.
%if time and data are available. 

Our approach is also related to methods for initialization of deep networks. %, in that our meta-learning mechanism aims to learn weights that can be 
In computer vision, models pretrained on large-scale image classification have been shown to learn effective features for a range of problems~\cite{decaf}. In contrast, our method explicitly optimizes the model for fast adaptability, allowing it to adapt to new tasks with only a few examples.
%a minuscule number of examples. %As discussed in Section~\ref{sec:intro}, 
Our method can also be viewed as explicitly maximizing sensitivity of new task losses to the model parameters.
%, so that small parameter changes can lead to large improvements in the loss. 
A number of prior works have explored sensitivity in deep networks, often in the context of initialization~\cite{orthogonal,forgetting}. Most of these works have considered good random initializations, though a number of papers have addressed data-dependent initializers~\cite{datadependentinit,weightnorm}, including learned initializations~\cite{husken,maclaurin}. In contrast, our method explicitly trains the parameters for sensitivity on a given task distribution, allowing for extremely efficient adaptation for problems such as $K$-shot learning and rapid reinforcement learning in only one or a few gradient steps.

\section{Experimental Evaluation}

\iffalse
\begin{figure*}
\setlength{\unitlength}{0.5\columnwidth}
\begin{picture}(1.99,0.8) \linethickness{0.5pt}
\put(0.0,-0.1){\includegraphics[width=0.67\columnwidth]{mse_5shot.png}}
\put(1.33,-0.1){\includegraphics[width=0.67\columnwidth]{mse_10shot.png}}
\put(2.67,-0.1){\includegraphics[width=0.67\columnwidth]{mse_20shot.png}}
\end{picture}
\caption{Quantitative sinusoid regression results showing test-time learning curves. 
Note that MAML continues to improve with additional gradient steps without overfitting to the extremely small dataset during meta-testing, and achieves a loss that is substantially lower than the baseline fine-tuning approach.
\label{fig:mamlquant}
\vspace{-0.5cm}
}
\end{figure*}
\fi

\begin{figure*}
\setlength{\unitlength}{0.5\columnwidth}
\begin{picture}(1.99,0.8) \linethickness{0.5pt}
\put(0.0,-0.0){\includegraphics[width=0.5\columnwidth]{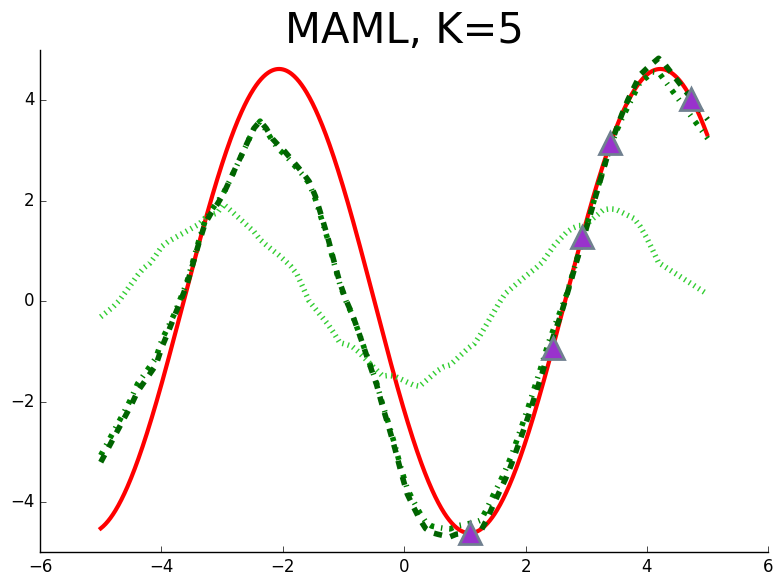}}
\put(0.95,-0.0){\includegraphics[width=0.5\columnwidth]{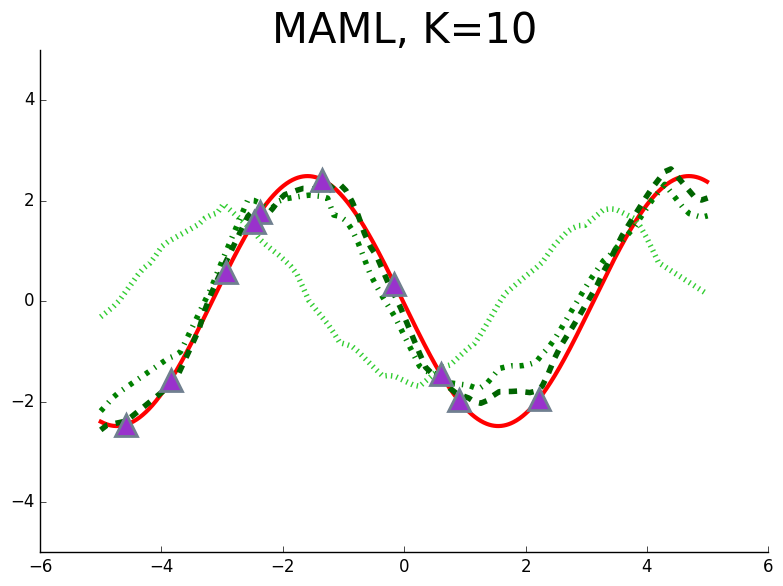}}

\put(2,0.05){\line(0,1){0.7}}

\put(2.05,-0.0){\includegraphics[width=0.5\columnwidth]{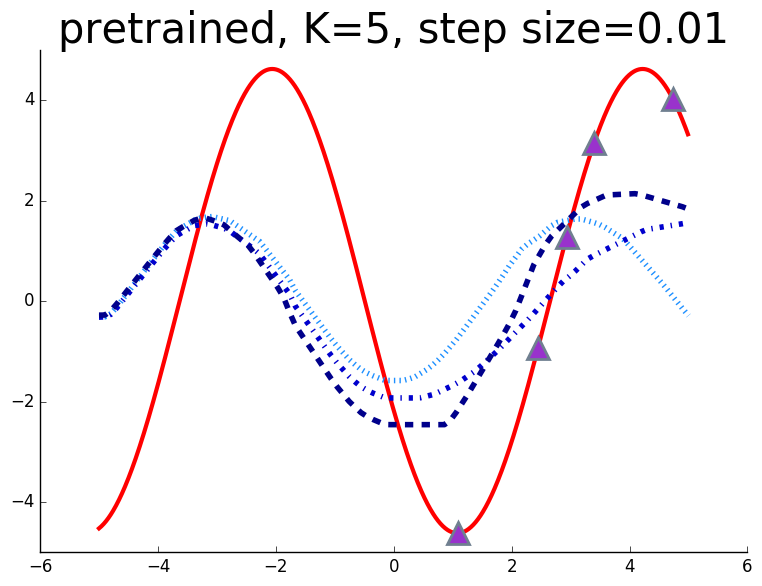}}
\put(3.0,-0.0){\includegraphics[width=0.49\columnwidth]{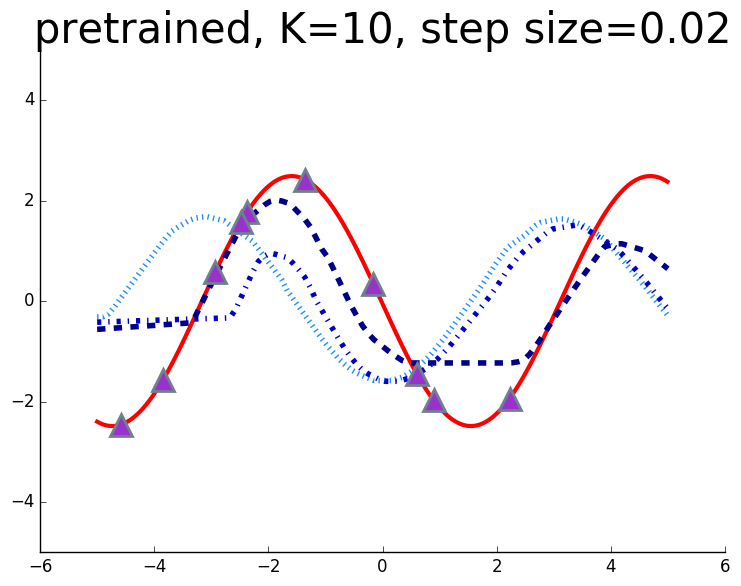}}
%\put(2.95,-0.1){(b)}

%\put(0.8,0.7){\includegraphics[height=0.1\columnwidth]{legend_maml.png}}
%\put(2.8,0.7){\includegraphics[height=0.1\columnwidth]{legend_pretrain.png}}

%\put(1.45,-0.1){\includegraphics[height=0.04\columnwidth]{legend_gt.png}}
%\put(2.03,-0.1){\includegraphics[height=0.04\columnwidth]{legend_gradpoints.png}}
\put(0.0,-0.1){\includegraphics[width=2\columnwidth]{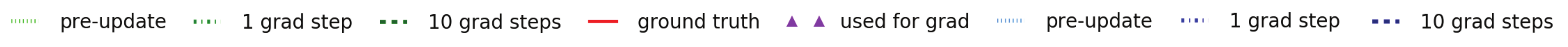}}
\end{picture}
\caption{Few-shot adaptation for the simple regression task. Left: Note that MAML is able to estimate parts of the curve where there are no datapoints, indicating that the model has learned about the periodic structure of sine waves. Right: Fine-tuning of a model pretrained on the same distribution of tasks without MAML, with a tuned step size. Due to the often contradictory outputs on the pre-training tasks, this model is unable to recover a suitable representation and fails to extrapolate from the small number of test-time samples.
\label{fig:mamlqual}
\vspace{-0.4cm}
}
\end{figure*}

\begin{figure}
\setlength{\unitlength}{0.5\columnwidth}
\begin{picture}(1.99,0.75) \linethickness{0.5pt}
%\put(0.0,-0.1){\includegraphics[width=0.67\columnwidth]{mse_5shot.png}}
\put(0.3,-0.15){\includegraphics[width=0.67\columnwidth]{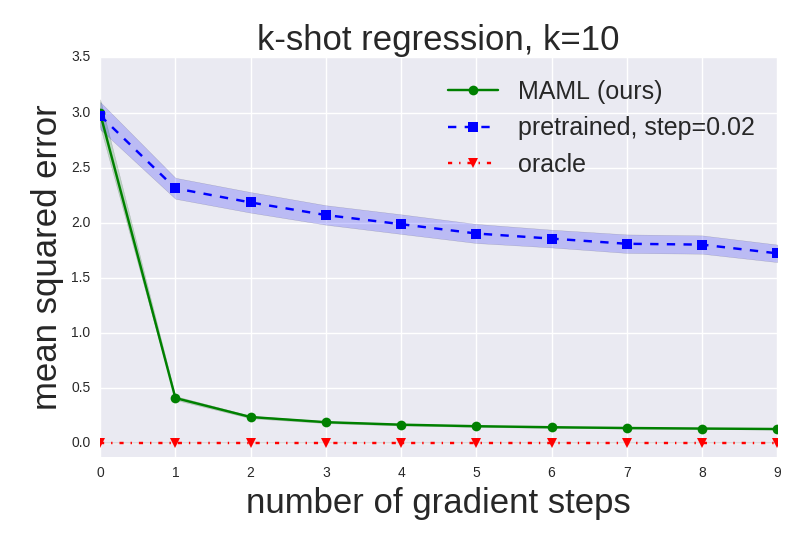}}
%\put(2.67,-0.1){\includegraphics[width=0.67\columnwidth]{mse_20shot.png}}
\end{picture}
\caption{Quantitative sinusoid regression results showing the learning curve at meta test-time. 
Note that MAML continues to improve with additional gradient steps without overfitting to the extremely small dataset during meta-testing, achieving a loss that is substantially lower than the baseline fine-tuning approach.
\label{fig:mamlquant}
\vspace{-0.5cm}
}
\end{figure}

\iffalse
\begin{table*}[t]
\vspace{-0.5cm}
\caption{Few-shot classification test accuracy on the MiniImagenet dataset by~\citet{hugo}. MAML achieves results that outperform state-of-the-art models. The $\pm$ shows $95\%$ confidence intervals over tasks. The evaluation of the baseline methods and matching networks is from~\citet{hugo}.}
\label{tbl:miniimagenet}
\begin{center}
{\footnotesize
\begin{tabular}{|l|c|c|c|c|}
\hline
\multicolumn{1}{|c}{} & \multicolumn{2}{|c|}{5-way Accuracy} \\
&  1-shot &  5-shot \\
\hline
fine-tuning baseline & $28.86 \pm 0.54\%$ & $49.79 \pm 0.79\%$ \\
\hline
nearest neighbor baseline & $41.08 \pm 0.70\%$ & $51.04 \pm 0.65\%$ \\
\hline
matching nets~\cite{matchingnets} & $43.56 \pm 0.84\%$ & $55.31 \pm 0.73\%$ \\
\hline
meta-learner LSTM~\cite{hugo} & $43.44 \pm 0.77\%$ & $60.60 \pm 0.71\%$  \\
\hline
\textbf{MAML (ours)}  & $\mathbf{47.70 \pm 1.79\%}$ & $\mathbf{62.63 \pm 0.94\%}$  \\
\hline
\end{tabular}
}
\end{center}
\vspace{-0.2cm}
\end{table*}
\fi

The goal of our experimental evaluation is to answer the following questions: (1) Can MAML enable fast learning of new tasks? (2) Can MAML be used for meta-learning in multiple different domains, including supervised regression, classification, and reinforcement learning? (3) Can a model learned with MAML continue to improve with additional gradient updates and/or examples?
% This 3rd question is essentially about sensitivity.

All of the meta-learning problems that we consider require some amount of adaptation to new tasks at test-time. When possible, we compare our results to an oracle that receives the identity of the task (which is a problem-dependent representation) as an additional input, as an upper bound on the performance of the model. All of the experiments were performed using TensorFlow~\cite{tensorflow}, which allows for automatic differentiation through the gradient update(s) during meta-learning. The code is available online\footnote{Code for the regression and supervised experiments is at \url{github.com/cbfinn/maml} and code for the RL experiments is at \url{github.com/cbfinn/maml_rl}}.
%The code for all of the experiments will be released.
%%SL.05.08: Commented out the above for the arxiv version (can put it back in once we have a link).

%Note that when the task is fully encoded in the input to the model, the task can generally be solved in zero-shots.
%%SL.1.30: "learner function" feels like a weird term to me, maybe you mean "learned model" or something like that?
%the model can simply memorize all of the tasks rather than learning to adapt. In this scenario, the learner would not be able to adapt to new tasks at test time. For this reason, we consider tasks in which the input to the learner function (i.e. the observation) does not include enough information about the task to solve it in zero shots (tasks which require some amount of learning).
% CF: maybe give an example here?
%%SL.1.30: I see what you're trying to say here, but as written, it's a little bit convoluted. An example would definitely help, but also you say "we consider tasks..." which in this section technically means that your method is only designed to handle those tasks -- that's not true. We specifically test our method on those tasks, but it can handle the other tasks just fine as well. Perhaps this entire paragraph would better fit in the experiments section?
% CF: okay, we can put it there.

\subsection{Regression} 

%%PA: how about in the text explicitly including the equation: y = A sin(2pi f x + phase) (+ noise?) ?
%%SL.02.24: won't fix (no space)
We start with a simple regression problem that illustrates the basic principles of MAML. Each task involves regressing from the input to the output of a sine wave, where the amplitude and phase of the sinusoid are varied between tasks. Thus, $p(\task)$ is continuous, where the amplitude varies within $[0.1,5.0]$ and the phase varies within $[0,\pi]$, and the input and output both have a dimensionality of $1$. During training and testing, datapoints $\inp$ are sampled uniformly from $[-5.0,5.0]$. The loss is the mean-squared error between the prediction $\learner(\inp)$ and true value. The regressor is a neural network model with $2$ hidden layers of size $40$ with ReLU nonlinearities. When training with MAML, we use one gradient update with $K=10$ examples with a fixed step size $\alpha=0.01$, and use Adam  as the meta-optimizer~\cite{adam}. The baselines are likewise trained with Adam. To evaluate performance, we fine-tune a single meta-learned model on varying numbers of $K$ examples, and compare performance to two baselines: (a) pretraining on all of the tasks, which entails training a network to regress to random sinusoid functions and then, at test-time, fine-tuning with gradient descent on the $K$ provided points, using an automatically tuned step size, and (b) an oracle which receives the true amplitude and phase as input. In Appendix~\ref{app:comparisons}, we show comparisons to additional multi-task and adaptation methods.
%comparisons to other baselines that perform worse than baseline (a) above.

We evaluate performance by fine-tuning the model learned by MAML and the pretrained model on $K=\{5,10,20\}$ datapoints. During fine-tuning, each gradient step is computed using the same $K$ datapoints. The qualitative results, shown in Figure~\ref{fig:mamlqual} and further expanded on in Appendix~\ref{app:sinequal} show that the learned model is able to quickly adapt with only $5$ datapoints, shown as purple triangles, whereas the model that is pretrained using standard supervised learning on all tasks is unable to adequately adapt with so few datapoints without catastrophic overfitting. Crucially, when the $K$ datapoints are all in one half of the input range, the model trained with MAML can still infer the amplitude and phase in the other half of the range, demonstrating that the MAML trained model $\learner$ has learned to model the periodic nature of the sine wave. Furthermore, we observe both in the qualitative and quantitative results (Figure~\ref{fig:mamlquant} and Appendix~\ref{app:sinequal}) that the model learned with MAML continues to improve with additional gradient steps, despite being trained for maximal performance after one gradient step. This improvement suggests that MAML optimizes the parameters such that they lie in a region that is amenable to fast adaptation and is sensitive to loss functions from $p(\task)$, as discussed in Section~\ref{sec:maml}, rather than overfitting to parameters $\theta$ that only improve after one step.

\subsection{Classification}
\label{sec:image_results}

\begin{table*}[t]
\vspace{-0.25cm}
\caption{Few-shot classification on held-out Omniglot characters (top) and the MiniImagenet test set (bottom). MAML achieves results that are comparable to or outperform  state-of-the-art convolutional and recurrent models. Siamese nets, matching nets, and the memory module approaches are all specific to classification, and are not directly applicable to regression or RL scenarios. The $\pm$ shows $95\%$ confidence intervals over tasks. Note that the Omniglot results may not be strictly comparable since the train/test splits used in the prior work were not available. The MiniImagenet evaluation of baseline methods and matching networks is from~\citet{hugo}.}
\label{tbl:omniglot}
\begin{center}
{\footnotesize
\begin{tabular}{|l|c|c|c|c|}
\hline
\multicolumn{1}{|c}{} & \multicolumn{2}{|c|}{5-way Accuracy} & \multicolumn{2}{|c|}{20-way Accuracy}\\
{ Omniglot~\citep{omniglot}}  &  1-shot &  5-shot &  1-shot &  5-shot\\
\hline
MANN, no conv ~\cite{mann} & $82.8\%$ & $94.9\%$ &-- & -- \\
\hline
\textbf{MAML, no conv (ours)} & $\mathbf{89.7 \pm 1.1}\%$ & $\mathbf{97.5 \pm 0.6}\%$ & -- & -- \\
\hline
\hline
Siamese nets~\cite{siameseoneshot} & $97.3\%$ & $98.4\%$ & $88.2\%$ & $97.0\%$ \\
\hline
matching nets~\cite{matchingnets} & $98.1\%$ & $98.9\%$ & $93.8\%$ & $98.5\%$ \\
\hline
neural statistician~\cite{neuralstatistician} & $98.1\%$ & $99.5\%$ & $93.2\%$ & $98.1\%$ \\
\hline
memory mod.~\cite{rareevents} & $98.4\%$ & $99.6\%$ & $95.0\%$ & $98.6\%$ \\
\hline
\textbf{MAML (ours)}  & $\mathbf{98.7 \pm 0.4\%}$ & $\mathbf{99.9 \pm 0.1\%}$ & $\mathbf{95.8 \pm 0.3\%}$ & $\mathbf{98.9 \pm 0.2}\%$ \\
\hline
 \multicolumn{5}{c}{} 
\end{tabular}
\begin{tabular}{|l|c|c|c|c|}
\hline
\multicolumn{1}{|c}{} & \multicolumn{2}{|c|}{5-way Accuracy} \\
MiniImagenet~\citep{hugo}  &  1-shot &  5-shot \\
\hline
fine-tuning baseline & $28.86 \pm 0.54\%$ & $49.79 \pm 0.79\%$ \\
\hline
nearest neighbor baseline & $41.08 \pm 0.70\%$ & $51.04 \pm 0.65\%$ \\
\hline
matching nets~\cite{matchingnets} & $43.56 \pm 0.84\%$ & $55.31 \pm 0.73\%$ \\
\hline
meta-learner LSTM~\cite{hugo} & $43.44 \pm 0.77\%$ & $60.60 \pm 0.71\%$  \\
\hline
\textbf{MAML, first order approx. (ours)}  & $\mathbf{48.07 \pm 1.75\%}$ & $\mathbf{63.15 \pm 0.91\%}$  \\
\hline
\textbf{MAML (ours)}  & $\mathbf{48.70 \pm 1.84\%}$ & $\mathbf{63.11 \pm 0.92\%}$  \\
\hline
\end{tabular}
}
\end{center}
\vspace{-0.35cm}
\end{table*}

To evaluate MAML in comparison to prior meta-learning and few-shot learning algorithms, we applied our method to few-shot image recognition on the Omniglot~\cite{omniglot} and MiniImagenet datasets. The Omniglot dataset consists of 20 instances of 1623 characters from 50 different alphabets. Each instance was drawn by a different person. The MiniImagenet dataset was proposed by~\citet{hugo}, and involves 64 training classes, 12 validation classes, and 24 test classes. The Omniglot and MiniImagenet image recognition tasks are the most common recently used few-shot learning benchmarks~\cite{matchingnets,mann,hugo}. We follow the experimental protocol proposed by~\citet{matchingnets}, which involves fast learning of $N$-way classification with 1 or 5 shots. The problem of $N$-way classification is set up as follows: select $N$ unseen classes, provide the model with $K$ different instances of each of the $N$ classes, and evaluate the model's ability to classify new instances within the $N$ classes. For Omniglot, we randomly select $1200$ characters for training, irrespective of alphabet, and use the remaining for testing. The Omniglot dataset is augmented with rotations by multiples of $90$ degrees, as proposed by~\citet{mann}.

Our model follows the same architecture as the embedding function used by~\citet{matchingnets}, which has 4 modules with a $3\times3$ convolutions and $64$ filters, followed by batch normalization~\cite{batchnorm}, a ReLU nonlinearity, and $2\times2$ max-pooling. The Omniglot images are downsampled to $28\times28$, so the dimensionality of the last hidden layer is $64$. As in the baseline classifier used by~\citet{matchingnets}, the last layer is fed into a softmax. For Omniglot, we used strided convolutions instead of max-pooling. For MiniImagenet, we used $32$ filters per layer to reduce overfitting, as done by~\cite{hugo}. 
%the last hidden layer of the MAML model is fed into a fully-connected layer followed by a softmax.
In order to also provide a fair comparison against memory-augmented neural networks~\cite{mann} and to test the flexibility of MAML, we also provide results for a non-convolutional network. For this, we use a network with $4$ hidden layers with sizes $256$, $128$, $64$, $64$, each including batch normalization and ReLU nonlinearities, followed by a linear layer and softmax. For all models, the loss function is the cross-entropy error between the predicted and true class. Additional hyperparameter details are included in Appendix~\ref{app:hyperclass}.

We present the results in Table~\ref{tbl:omniglot}. The convolutional model learned by MAML compares well to the state-of-the-art results on this task, narrowly outperforming the prior methods. Some of these existing methods, such as matching networks, Siamese networks, and memory models are designed with few-shot classification in mind, and are not readily applicable to domains such as reinforcement learning. Additionally, the model learned with MAML uses fewer overall parameters compared to matching networks and the meta-learner LSTM, since the algorithm does not introduce any additional parameters beyond the weights of the classifier itself. Compared to these prior methods, memory-augmented neural networks~\cite{mann} specifically, and recurrent meta-learning models in general, represent a more broadly applicable class of methods that, like MAML, can be used for other tasks such as reinforcement learning~\cite{rl2,learningrl}. However, as shown in the comparison, MAML significantly outperforms memory-augmented networks and the meta-learner LSTM on 5-way Omniglot and MiniImagenet classification, both in the $1$-shot and $5$-shot case.

A significant computational expense in MAML comes from the use of second derivatives when backpropagating the meta-gradient through the gradient operator in the meta-objective (see Equation~(\ref{eq:metaupdate})). On MiniImagenet, we show a comparison to a first-order approximation of MAML, where these second derivatives are omitted. Note that the resulting method still computes the meta-gradient at the post-update parameter values $\theta_i'$, which provides for effective meta-learning. Surprisingly however, the performance of this method is nearly the same as that obtained with full second derivatives, suggesting that most of the improvement in MAML comes from the gradients of the objective at the post-update parameter values, rather than the second order updates from differentiating through the gradient update. Past work has observed that ReLU neural networks are locally almost linear~\citep{linear}, which suggests that second derivatives may be close to zero in most cases, partially explaining the good performance of the first-order approximation. This approximation removes the need for computing Hessian-vector products in an additional backward pass, which we found led to roughly $33\%$ speed-up in network computation.

\subsection{Reinforcement Learning}

\iffalse
\begin{figure}
\setlength{\unitlength}{0.5\columnwidth}
\begin{picture}(1.99,0.71) \linethickness{0.5pt}
\put(-0.01,0.0){\includegraphics[width=0.51\columnwidth]{point_results.png}}
\put(1.03,-0.1){\includegraphics[width=0.48\columnwidth]{pretrain_paths_viz.png}}
\put(1.03,0.36){\includegraphics[width=0.48\columnwidth]{maml_paths_viz.png}}
\end{picture}
\caption{Left: quantitative results from 2D navigation task, Right: qualitative comparison between model learned with MAML and with fine-tuning from a pretrained network.
\label{fig:2d}
\vspace{-0.5cm}
}
\end{figure}
\fi

\begin{figure}
\setlength{\unitlength}{0.5\columnwidth}
\begin{picture}(1.99,1.05) \linethickness{0.5pt}
\put(0.4,0.4){\includegraphics[width=0.51\columnwidth]{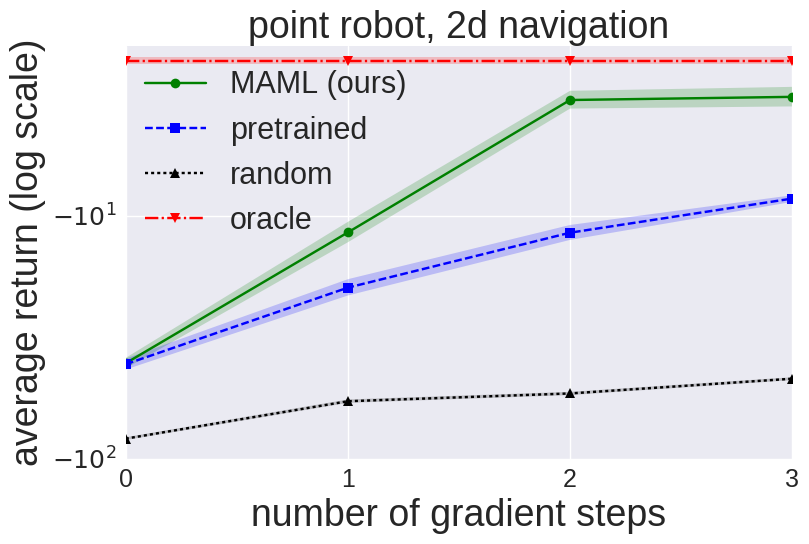}}
\put(0.99,-0.1){\includegraphics[width=0.51\columnwidth]{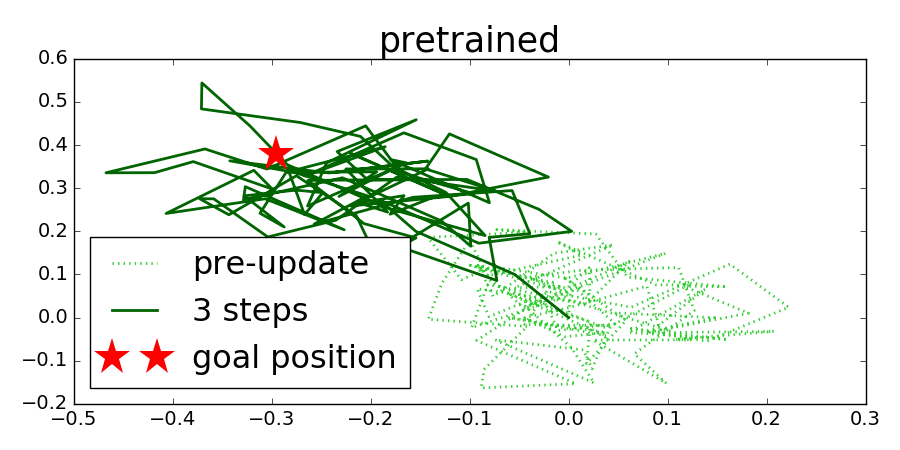}}
\put(0.0,-0.1){\includegraphics[width=0.51\columnwidth]{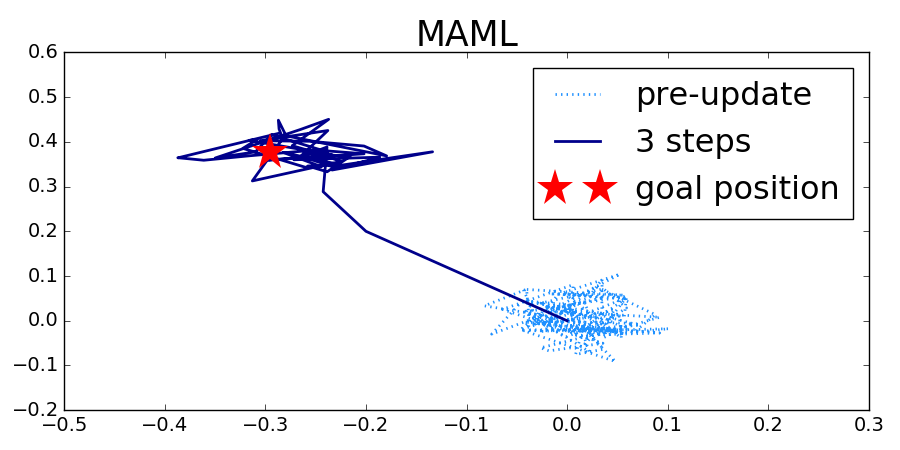}}
\end{picture}
\caption{Top: quantitative results from 2D navigation task, Bottom: qualitative comparison between model learned with MAML and with fine-tuning from a pretrained network.
\label{fig:2d}
\vspace{-0.6cm}
}
\end{figure}

\begin{figure*}
\setlength{\unitlength}{0.5\columnwidth}
\begin{picture}(1.99,0.56) \linethickness{0.5pt}

\put(2.72,-0.1){\includegraphics[width=0.49\columnwidth]{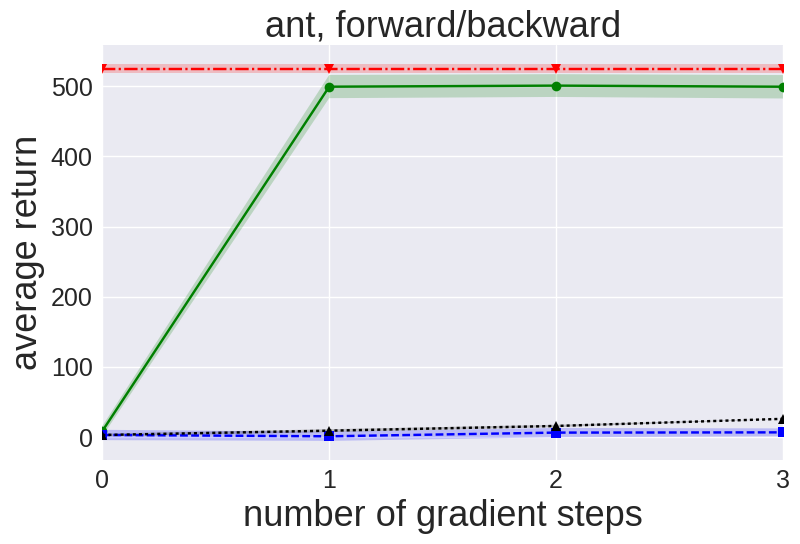}}
\put(1.80,-0.1){\includegraphics[width=0.49\columnwidth]{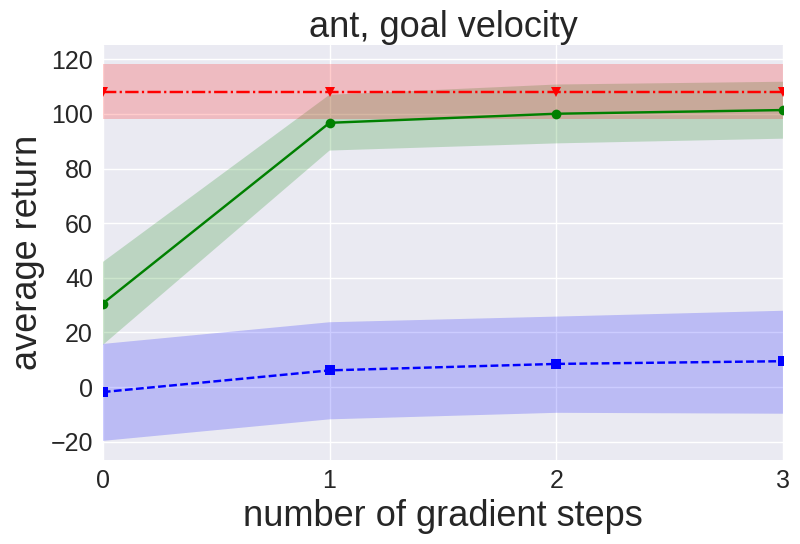}}
\put(0.885,-0.1){\includegraphics[width=0.49\columnwidth]{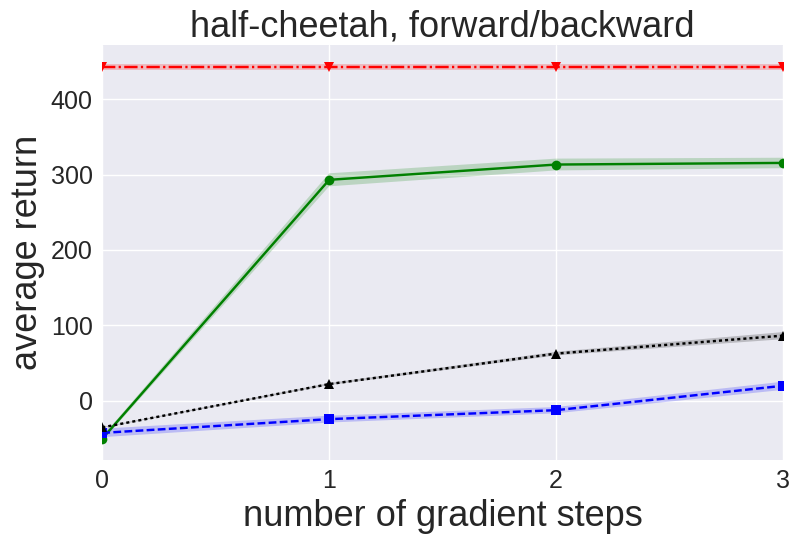}}
\put(-0.03,-0.1){\includegraphics[width=0.49\columnwidth]{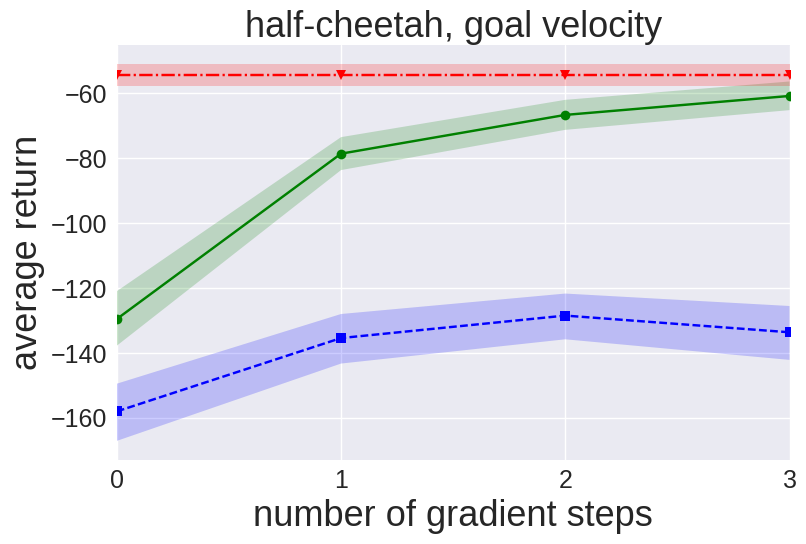}}

\put(3.28,0.1){\includegraphics[width=0.2\columnwidth]{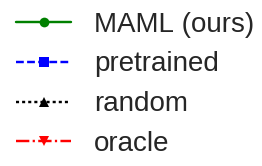}}

\put(3.69, 0.225){\includegraphics[width=0.22\columnwidth]{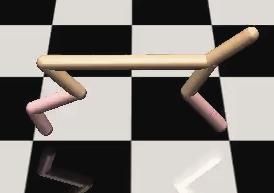}}
\put(3.69,-0.07){\includegraphics[width=0.22\columnwidth]{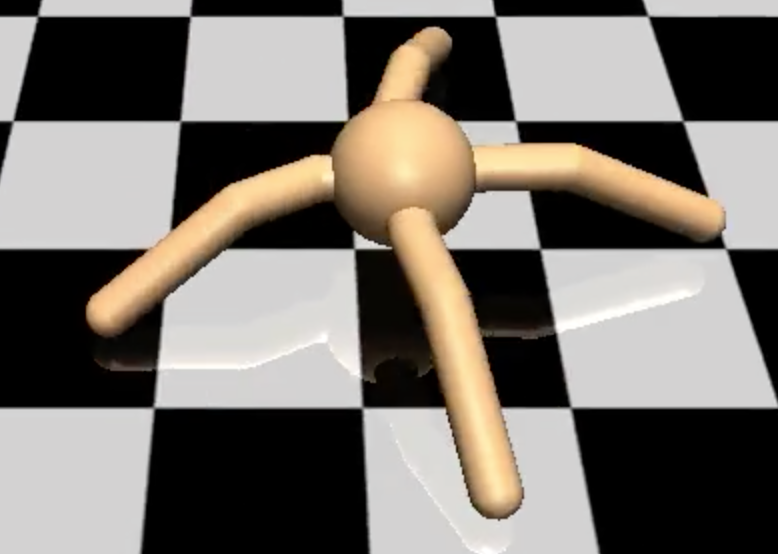}}

\end{picture}
\caption{Reinforcement learning results for the half-cheetah and ant locomotion tasks, with the tasks shown on the far right. Each gradient step requires additional samples from the environment, unlike the supervised learning tasks. The results show that MAML can adapt to new goal velocities and directions substantially faster than conventional pretraining or random initialization, achieving good performs in just two or three gradient steps. We exclude the goal velocity, random baseline curves, since the returns are much worse ($<-200$ for cheetah and $<-25$ for ant).
\label{fig:loco}
\vspace{-0.5cm}
}
\end{figure*}

\iffalse
\begin{figure}
\setlength{\unitlength}{0.5\columnwidth}
\begin{picture}(1.99,0.65) \linethickness{0.5pt}
\put(0.0,-0.1){\includegraphics[height=0.35\columnwidth]{cheetahshot.png}}
\put(1,-0.1){\includegraphics[width=0.49\columnwidth]{antshot.png}}
\end{picture}
\caption{Illustration of the locomotion environments.
\label{fig:tasks}
\vspace{-0.5cm}
}
\end{figure}
\fi

To evaluate MAML on reinforcement learning problems, we constructed several sets of tasks based off of the simulated continuous control environments in the rllab benchmark suite~\cite{benchmarking}. We discuss the individual domains below. In all of the domains, the model trained by MAML is a neural network policy with two hidden layers of size $100$, with ReLU nonlinearities. The gradient updates are computed using vanilla policy gradient (REINFORCE)~\cite{reinforce}, and we use trust-region policy optimization (TRPO) as the meta-optimizer~\cite{trpo}. In order to avoid computing third derivatives, we use finite differences to compute the Hessian-vector products for TRPO. For both learning and meta-learning updates, we use the standard linear feature baseline proposed by~\citet{benchmarking}, which is fitted separately at each iteration for each sampled task in the batch. We compare to three baseline models: (a) pretraining one policy on all of the tasks and then fine-tuning, (b) training a policy from randomly initialized weights, and (c) an oracle policy which receives the parameters of the task as input, which for the tasks below corresponds to a goal position, goal direction, or goal velocity for the agent. The baseline models of (a) and (b) are fine-tuned with gradient descent with a manually tuned step size. Videos of the learned policies can be viewed at \mbox{\url{sites.google.com/view/maml}}
%%SL.02.23: I usually leave the period off the end of the url, in case someone tries to copy & paste and copies the period, and then wonders why it doesn't work

%CF: should we state somewhere that vanilla policy gradient is notoriously sample inefficient, in case we get a reviewer who doens't know anything about reinforcement learning?

\vspace{-0.1cm}
\noindent {\bf 2D Navigation.}
In our first meta-RL experiment, we study a set of tasks where a point agent must move to different goal positions in 2D, randomly chosen for each task within a unit square. The observation is the current 2D position, and actions correspond to velocity commands clipped to be in the range $[-0.1,0.1]$. The reward is the negative squared distance to the goal, and episodes terminate when the agent is within $0.01$ of the goal or at the horizon of $H=100$. The policy was trained with MAML to maximize performance after $1$ policy gradient update using $20$ trajectories. Additional hyperparameter settings for this problem and the following RL problems are in Appendix~\ref{app:hyperrl}.
In our evaluation, we compare adaptation to a new task with up to 4 gradient updates, each with $40$ samples. The results in Figure~\ref{fig:2d} show the adaptation performance of models that are initialized with MAML, conventional pretraining on the same set of tasks, random initialization, and an oracle policy that receives the goal position as input. The results show that MAML can learn a model that adapts much more quickly in a single gradient update, and furthermore continues to improve with additional updates.

%\subsubsection{Goal velocity}

\vspace{-0.1cm}
\noindent {\bf Locomotion.}
To study how well MAML can scale to more complex deep RL problems, we also study adaptation on high-dimensional locomotion tasks with the MuJoCo simulator ~\cite{mujoco}. The tasks require two simulated robots -- a planar cheetah and a 3D quadruped (the ``ant'') -- to run in a particular direction or at a particular velocity. In the goal velocity experiments, the reward is the negative absolute value between the current velocity of the agent and a goal, which is chosen uniformly at random between $0.0$ and $2.0$ for the cheetah and between $0.0$ and $3.0$ for the ant. In the goal direction experiments, the reward is the magnitude of the velocity in either the forward or backward direction, chosen at random for each task in $p(\task)$. The horizon is $H=200$, with $20$ rollouts per gradient step for all problems except the ant forward/backward task, which used $40$ rollouts per step. The results in Figure~\ref{fig:loco} show that MAML learns a model that can quickly adapt its velocity and direction with even just a single gradient update, and continues to improve with more gradient steps. The results also show that, on these challenging tasks, the MAML initialization substantially outperforms random initialization and pretraining. In fact, pretraining is in some cases worse than random initialization, a fact observed in prior RL work~\cite{actormimic}.
%%SL.02.24: describe alive bonus in appendix (it's a pretty standard trick), I left it out here. Otherwise, I did my best to summarize everything you had but in a more condensed format.

\iffalse
\begin{table*}[t]
\caption{MAML for meta-RL on 2D point robot - average return \todo{make this a line plot with error bars.}}
\label{tbl:pointmass}
\begin{center}
\begin{tabular}{c|c|c|c|c}
\hline
num. gradient updates & 0 & 1   & 2 & 3\\
\hline
random init. & $-82.66$ & $-57.96$ & $-53.84$ & $-46.80$ \\
\hline
init from all tasks & $-40.81$ & $-19.73$ & $-11.72$ & $-8.47$ \\
%\hline
%MAML, masking (ours) & $-42.42$ & $-13.90$ & $-5.17$ & $-3.18$ \\
\hline
MAML (ours) & $-40.41$ & $-11.68$ & $-3.33$ & $-3.23$ \\
\hline
\hline
oracle & -2.29 & n/a & n/a & n/a \\
\hline
\end{tabular}
\end{center}
\end{table*}
\fi

\section{Discussion and Future Work}

We introduced a meta-learning method based on learning easily adaptable model parameters through gradient descent. Our approach has a number of benefits. It is simple and does not introduce any learned parameters for meta-learning. It can be combined with any model representation that is amenable to gradient-based training, and any differentiable objective, including classification, regression, and reinforcement learning. Lastly, since our method merely produces a weight initialization, adaptation can be performed with any amount of data and any number of gradient steps, though we demonstrate state-of-the-art results on classification with only one or five examples per class. We also show that our method can adapt an RL agent using policy gradients and a very modest amount of experience.

Reusing knowledge from past tasks may be a crucial ingredient in making high-capacity scalable models, such as deep neural networks, amenable to fast training with small datasets. We believe that this work is one step toward a simple and general-purpose meta-learning technique that can be applied to any problem and any model. Further research in this area can make multitask initialization a standard ingredient in deep learning and reinforcement learning.

% Acknowledgements should only appear in the accepted version.
\section*{Acknowledgements} The authors would like to thank Xi Chen and Trevor Darrell for helpful discussions, Yan Duan and Alex Lee for technical advice, Nikhil Mishra, Haoran Tang, and Greg Kahn for feedback on an early draft of the paper, and the anonymous reviewers for their comments. This work was supported in part by an ONR PECASE award and an NSF GRFP award.

%\textbf{Do not} include acknowledgements in the initial version of
%the paper submitted for blind review.

%If a paper is accepted, the final camera-ready version can (and
%probably should) include acknowledgements. In this case, please
%place such acknowledgements in an unnumbered section at the
%end of the paper. Typically, this will include thanks to reviewers
%who gave useful comments, to colleagues who contributed to the ideas,
%and to funding agencies and corporate sponsors that provided financial
%support.

%\clearpage
\bibliography{references}
\bibliographystyle{icml2016}

\clearpage

\appendix

\section{Additional Experiment Details}
\label{app:hyper}

In this section, we provide additional details of the experimental set-up and hyperparameters.

\subsection{Classification}
\label{app:hyperclass}

%For N-way classification, the model learned by MAML was trained for N-way, 1-shot classification and evaluated on 1-shot and 5-shot evaluation.
% not true for miniimagenet...
For N-way, K-shot classification, each gradient is computed using a batch size of $NK$ examples.
%For 5-shot evaluation, we simply increased the batch size used for each gradient update from $N$ examples to $5N$ examples. 
For Omniglot, the 5-way convolutional and non-convolutional MAML models were each trained with $1$ gradient step with step size $\alpha=0.4$ and a meta batch-size of $32$ tasks. The network was evaluated using $3$ gradient steps with the same step size $\alpha=0.4$. The 20-way convolutional MAML model was trained and evaluated with $5$ gradient steps with step size $\alpha=0.1$. During training, the meta batch-size was set to $16$ tasks. For MiniImagenet, both models were trained using $5$ gradient steps of size $\alpha=0.01$, and evaluated using $10$ gradient steps at test time. Following~\citet{hugo}, $15$ examples per class were used for evaluating the post-update meta-gradient. We used a meta batch-size of $4$ and $2$ tasks for $1$-shot and $5$-shot training respectively.
All models were trained for $60000$ iterations on a single NVIDIA Pascal Titan X GPU. 

\subsection{Reinforcement Learning}
\label{app:hyperrl}

In all reinforcement learning experiments, the MAML policy was trained using a single gradient step with $\alpha=0.1$. During evaluation, we found that halving the learning rate after the first gradient step produced superior performance. Thus, the step size during adaptation was set to $\alpha=0.1$ for the first step, and $\alpha=0.05$ for all future steps. The step sizes for the baseline methods were manually tuned for each domain. In the 2D navigation, we used a meta batch size of $20$; in the locomotion problems, we used a meta batch size of $40$ tasks. The MAML models were trained for up to $500$ meta-iterations, and the model with the best average return during training was used for evaluation. For the ant goal velocity task, we added a positive reward bonus at each timestep to prevent the ant from ending the episode.

\section{Additional Sinusoid Results}
\label{app:sinequal}

\begin{figure*}
\setlength{\unitlength}{0.5\columnwidth}
\begin{picture}(1.99,0.8) \linethickness{0.5pt}
\put(0.0,-0.1){\includegraphics[width=0.67\columnwidth]{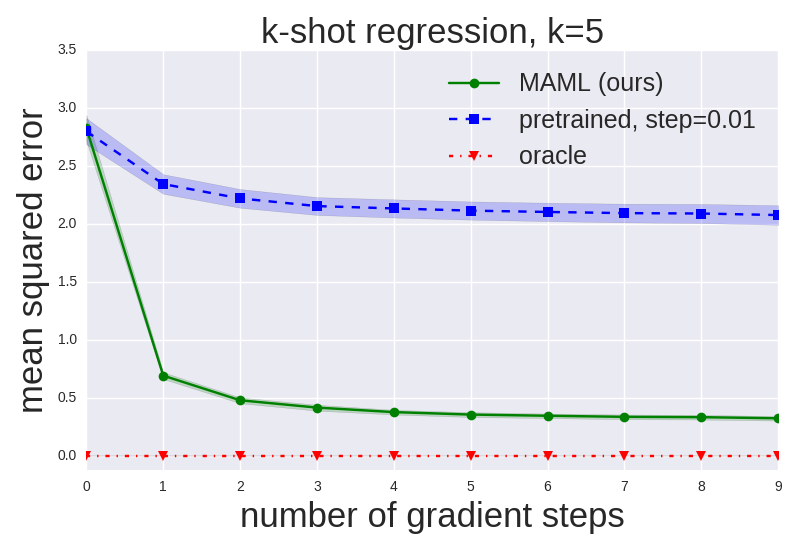}}
\put(1.33,-0.1){\includegraphics[width=0.67\columnwidth]{mse_10shot.png}}
\put(2.67,-0.1){\includegraphics[width=0.67\columnwidth]{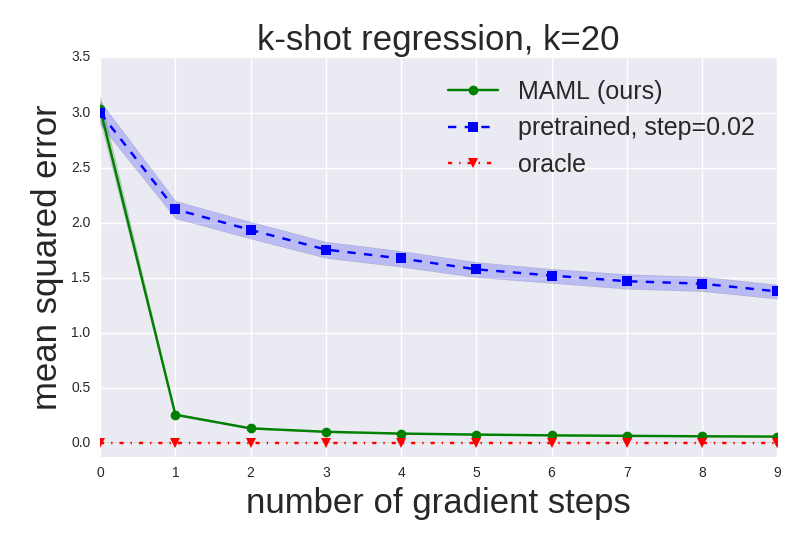}}
\end{picture}
\caption{Quantitative sinusoid regression results showing test-time learning curves with varying numbers of $K$ test-time samples. Each gradient step is computed using the same $K$ examples.
%The qualitative results show the learned function, while the quantitative results below show test-time learning curves with different numbers of $K$ test-time samples. 
Note that MAML continues to improve with additional gradient steps without overfitting to the extremely small dataset during meta-testing, and achieves a loss that is substantially lower than the baseline fine-tuning approach.
\label{fig:sinequant}
}
\end{figure*}

\begin{figure*}
\setlength{\unitlength}{0.5\columnwidth}
\begin{picture}(1.99,4.8) \linethickness{0.5pt}
\put(0.0,0.0){\includegraphics[width=0.5\columnwidth]{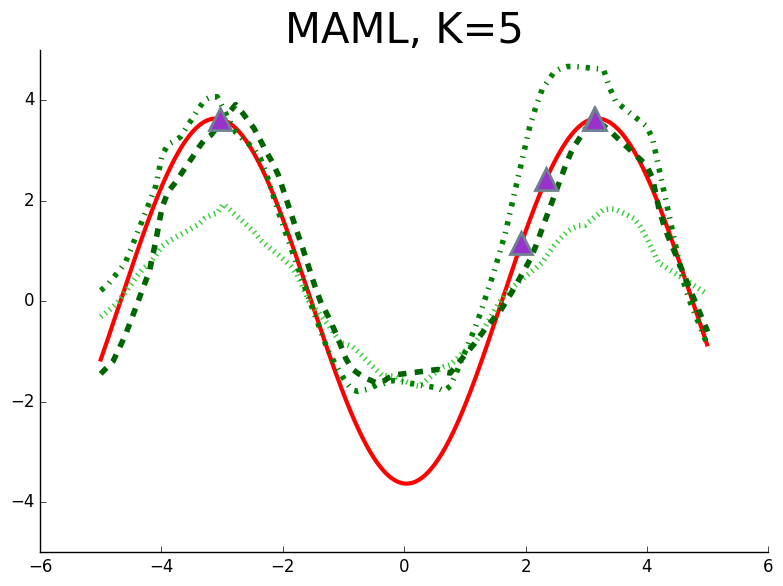}}
\put(0.0,0.8){\includegraphics[width=0.5\columnwidth]{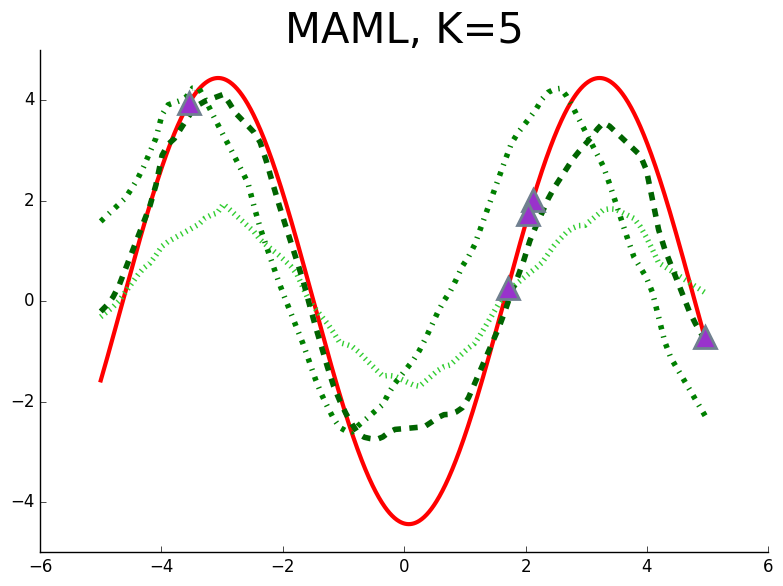}}
\put(0.0,1.6){\includegraphics[width=0.5\columnwidth]{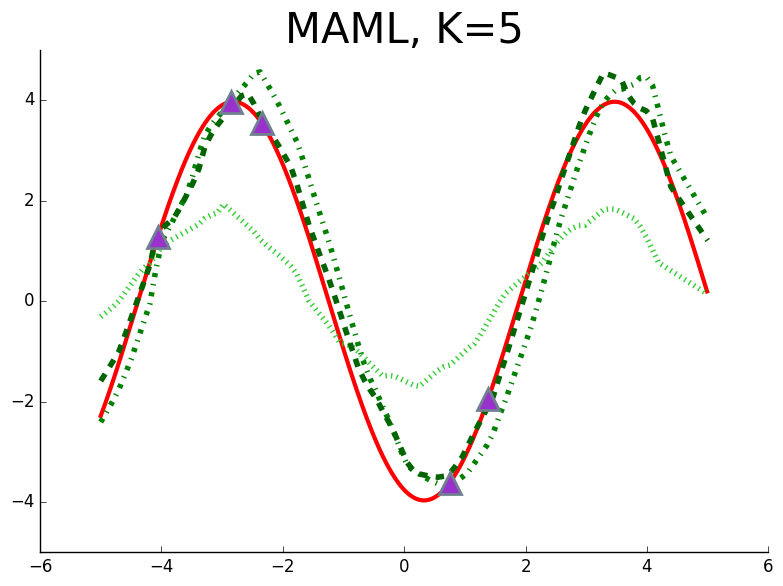}}
\put(0.0,2.4){\includegraphics[width=0.5\columnwidth]{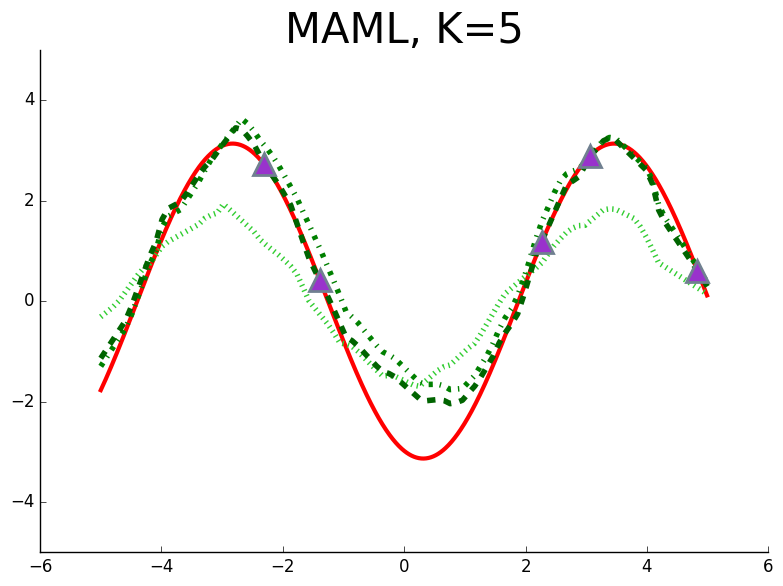}}
\put(0.0,3.2){\includegraphics[width=0.5\columnwidth]{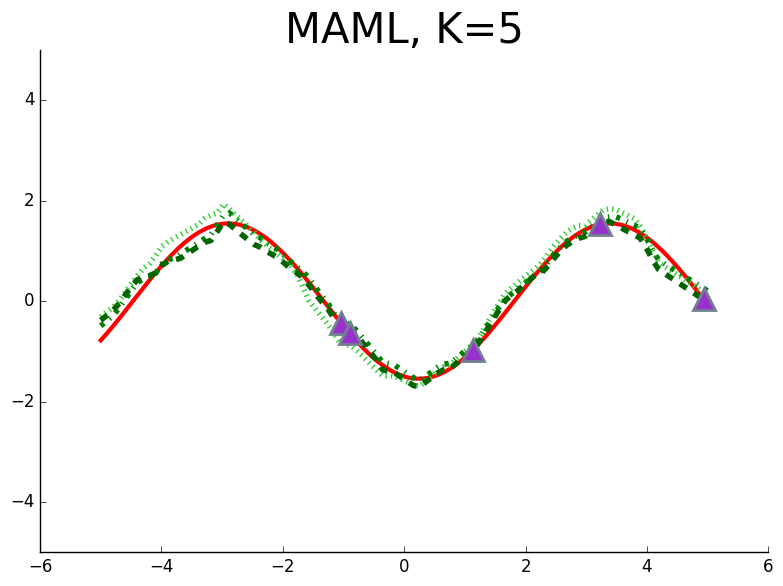}}
\put(0.0,4.0){\includegraphics[width=0.5\columnwidth]{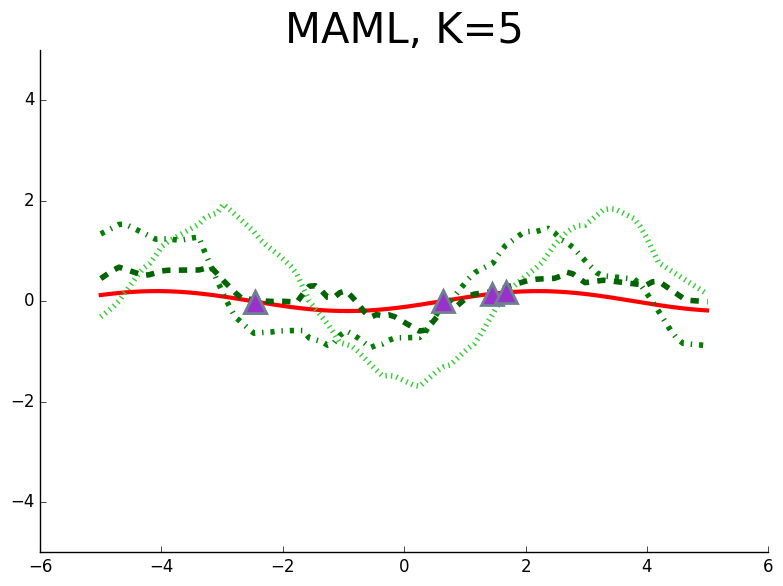}}

\put(1.0,0.0){\includegraphics[width=0.5\columnwidth]{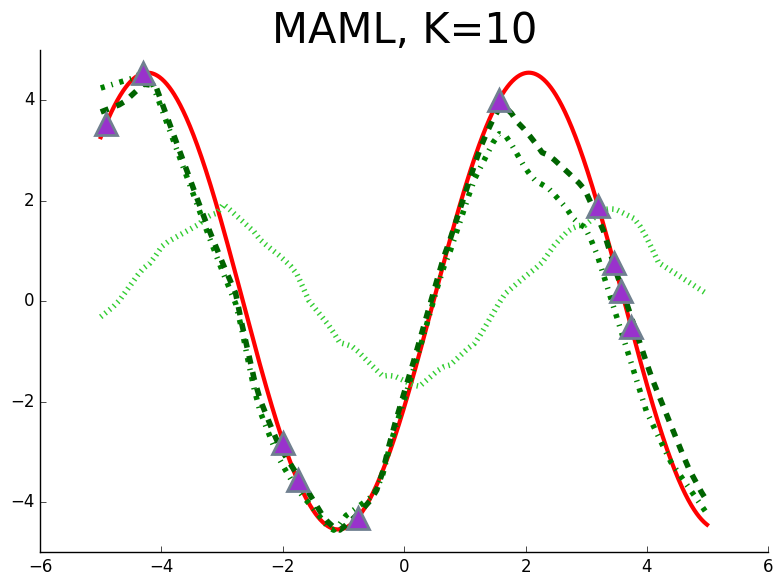}}
\put(1.0,0.8){\includegraphics[width=0.5\columnwidth]{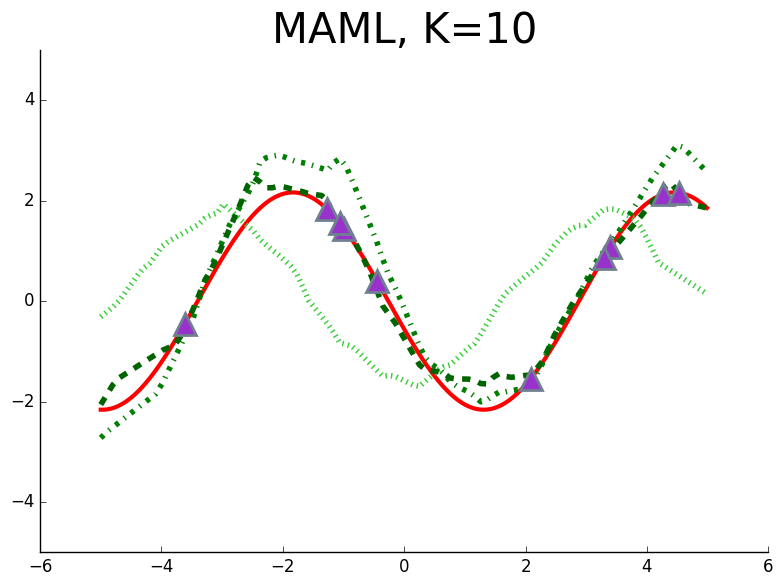}}
\put(1.0,1.6){\includegraphics[width=0.5\columnwidth]{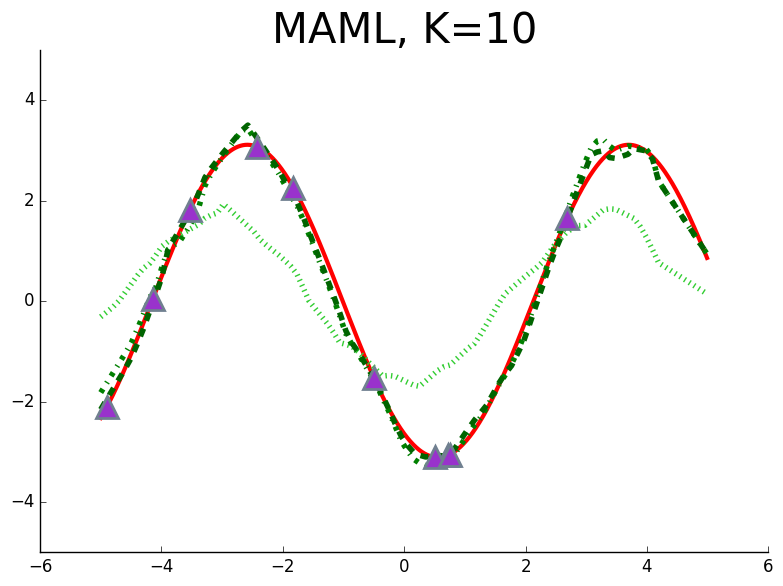}}
\put(1.0,2.4){\includegraphics[width=0.5\columnwidth]{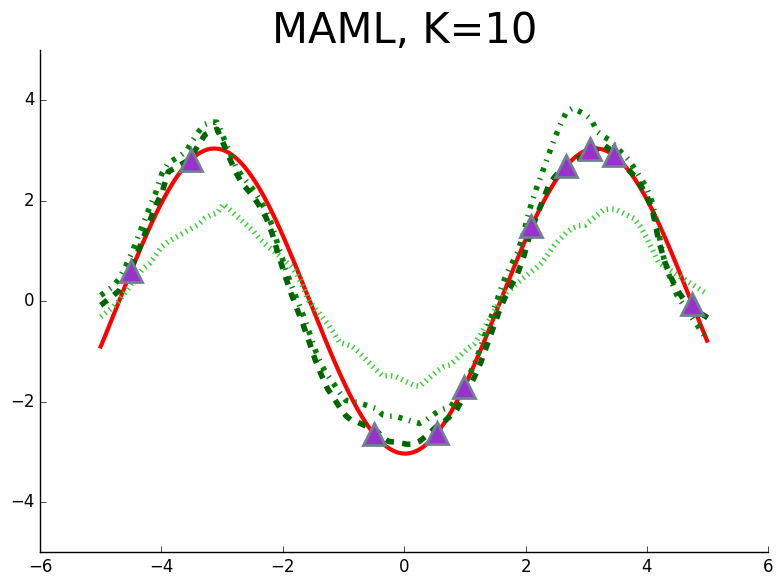}}
\put(1.0,3.2){\includegraphics[width=0.5\columnwidth]{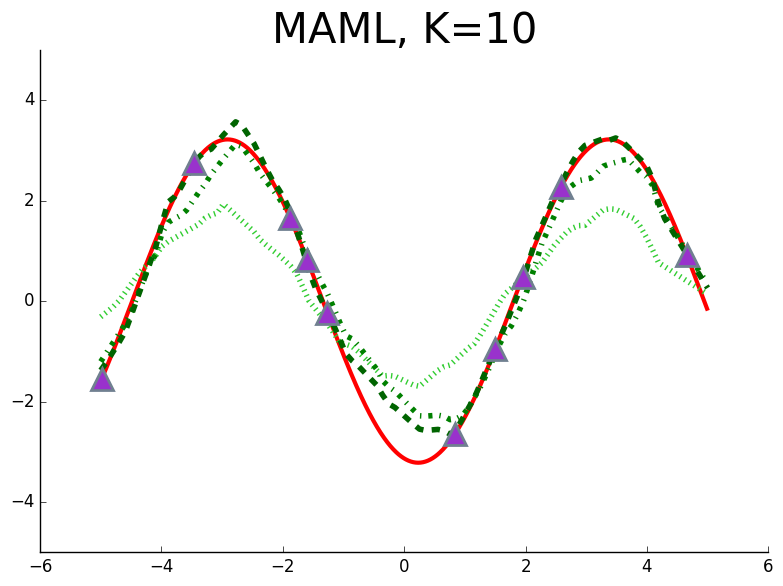}}
\put(1.0,4.0){\includegraphics[width=0.5\columnwidth]{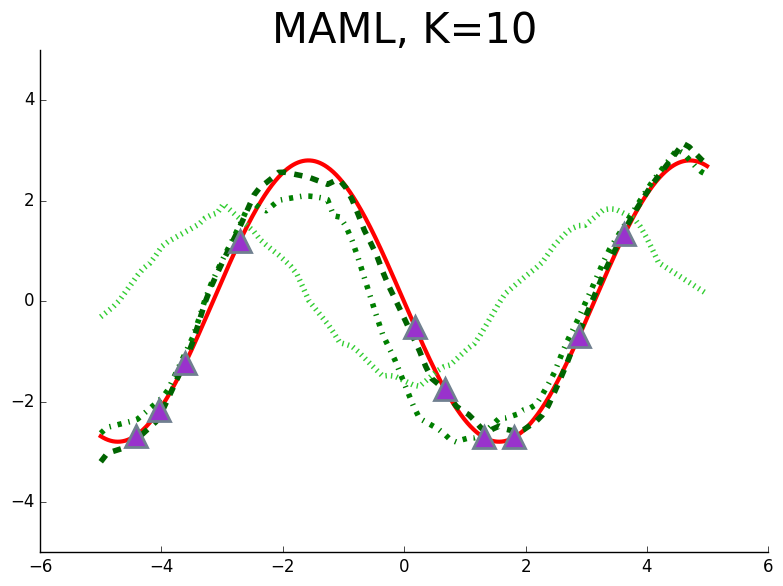}}

\put(2.0,0.0){\includegraphics[width=0.5\columnwidth]{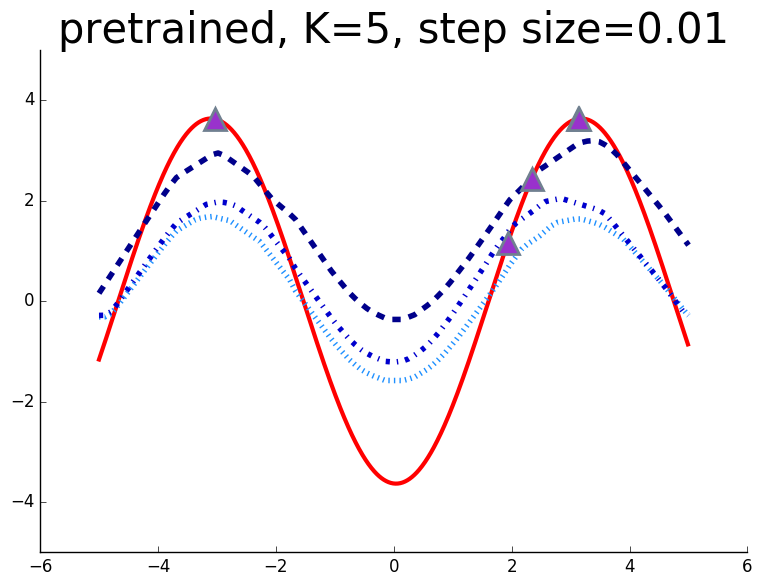}}
\put(2.0,0.8){\includegraphics[width=0.5\columnwidth]{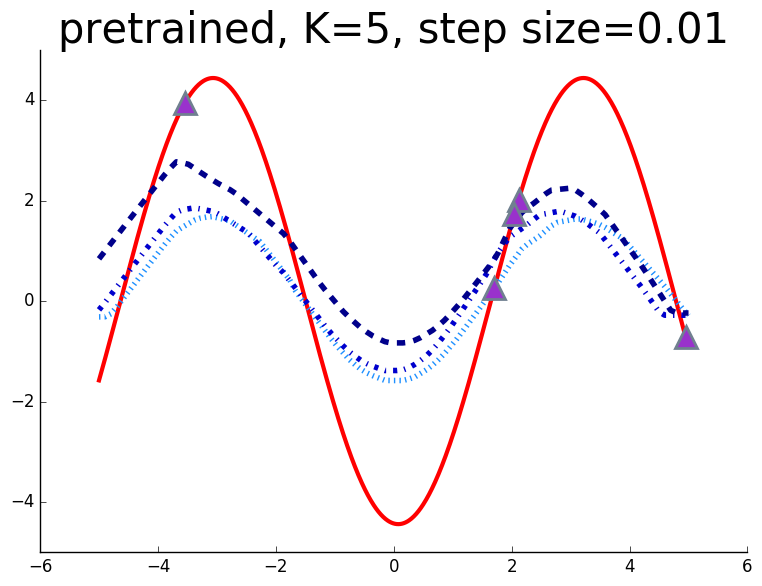}}
\put(2.0,1.6){\includegraphics[width=0.5\columnwidth]{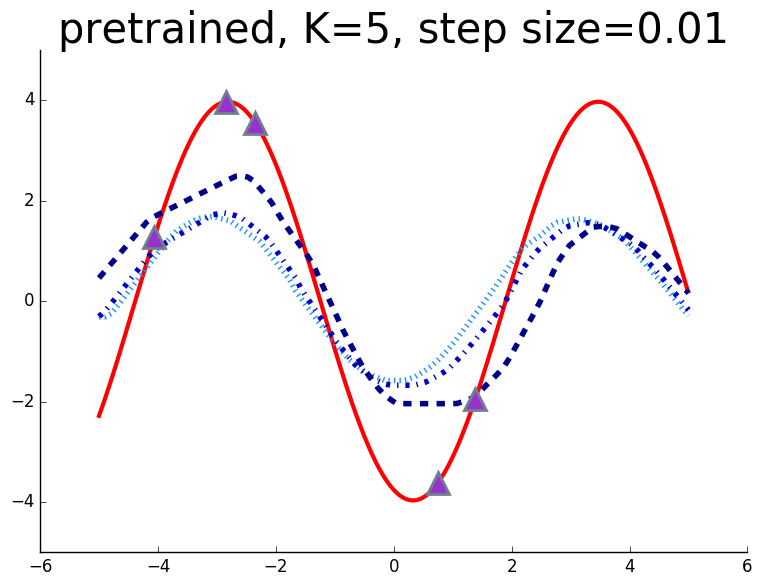}}
\put(2.0,2.4){\includegraphics[width=0.5\columnwidth]{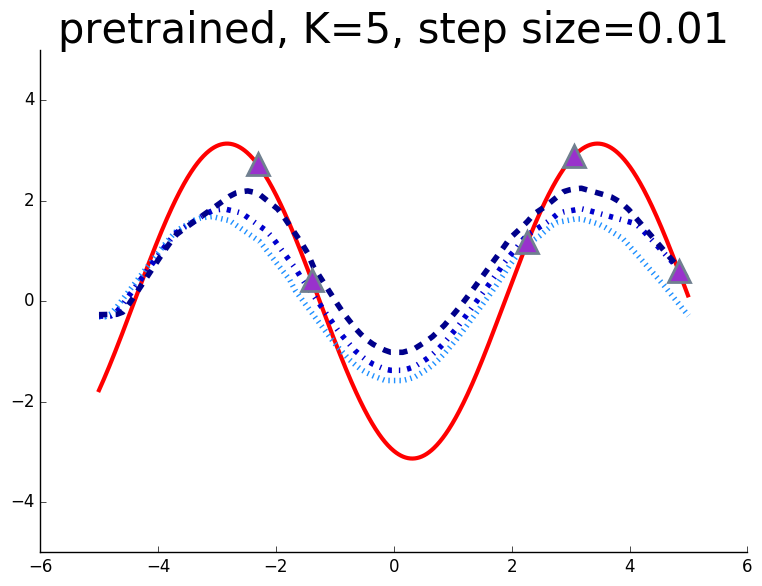}}
\put(2.0,3.2){\includegraphics[width=0.5\columnwidth]{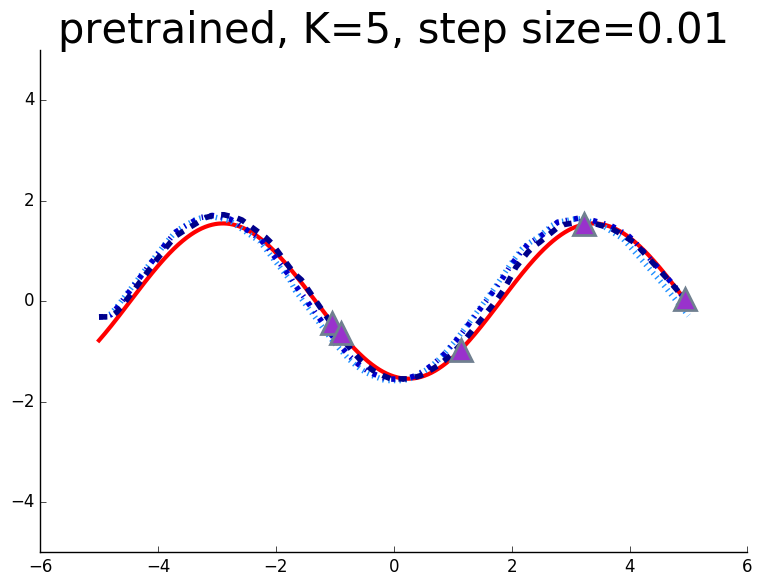}}
\put(2.0,4.0){\includegraphics[width=0.5\columnwidth]{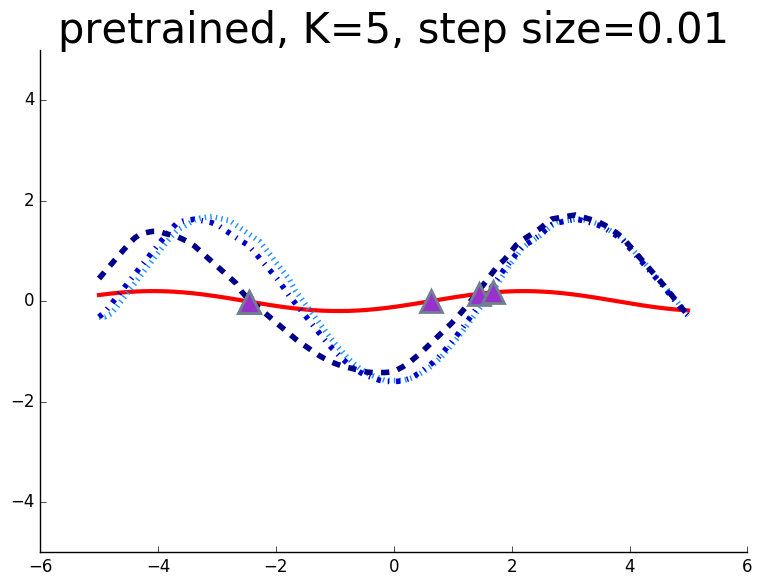}}

\put(3.0,0.0){\includegraphics[width=0.5\columnwidth]{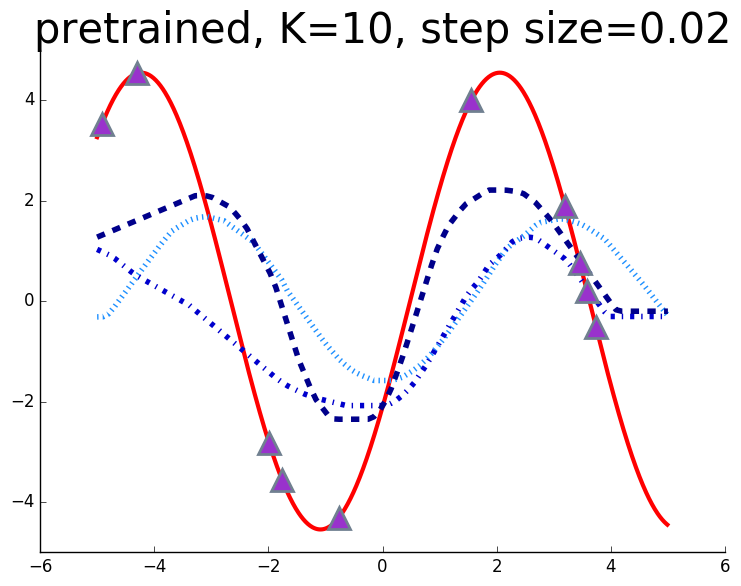}}
\put(3.0,0.8){\includegraphics[width=0.5\columnwidth]{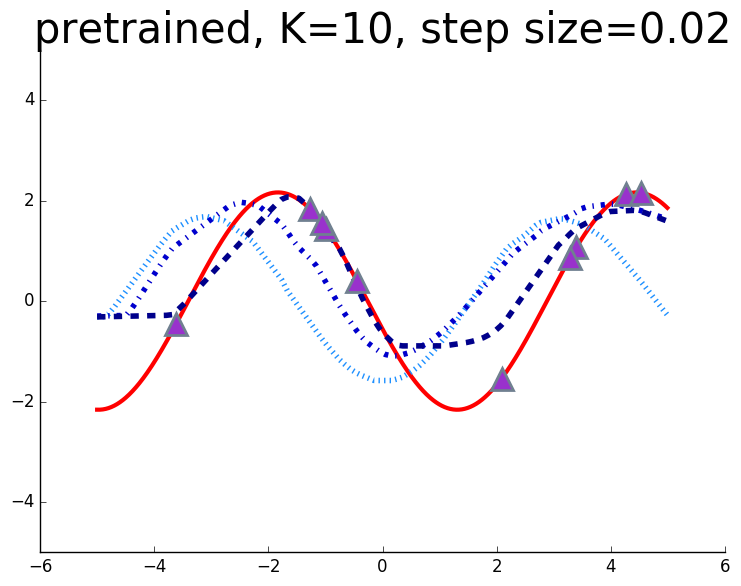}}
\put(3.0,1.6){\includegraphics[width=0.5\columnwidth]{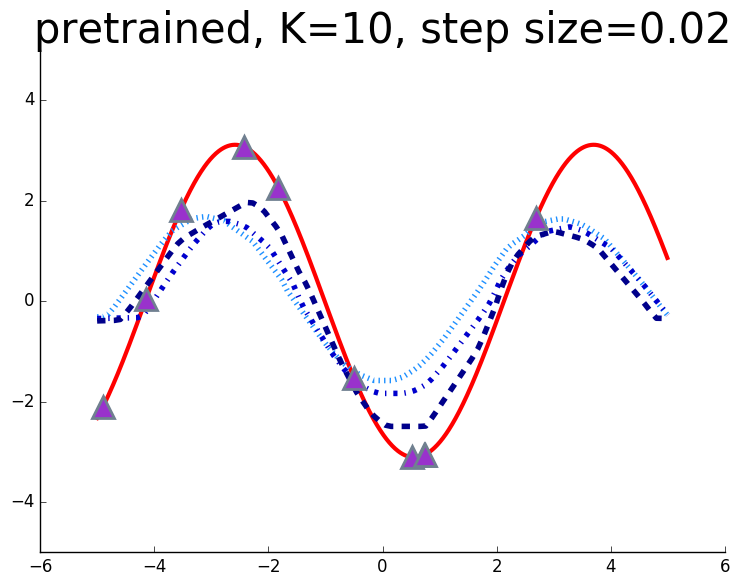}}
\put(3.0,2.4){\includegraphics[width=0.5\columnwidth]{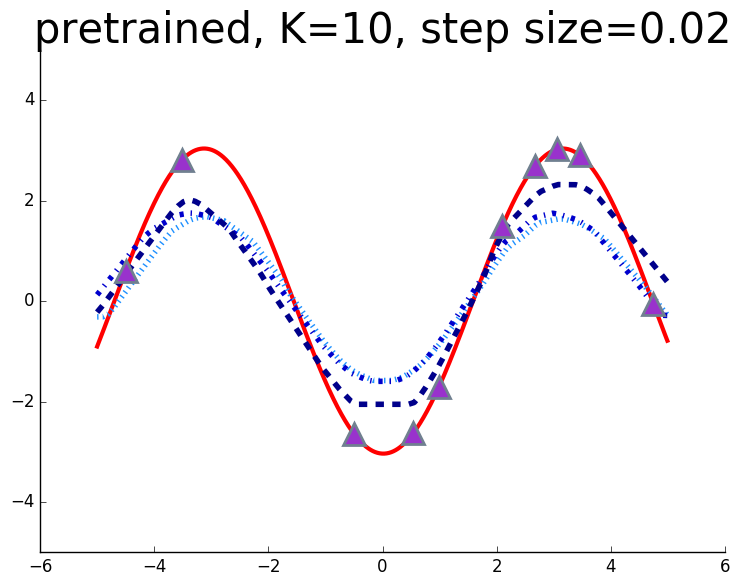}}
\put(3.0,3.2){\includegraphics[width=0.5\columnwidth]{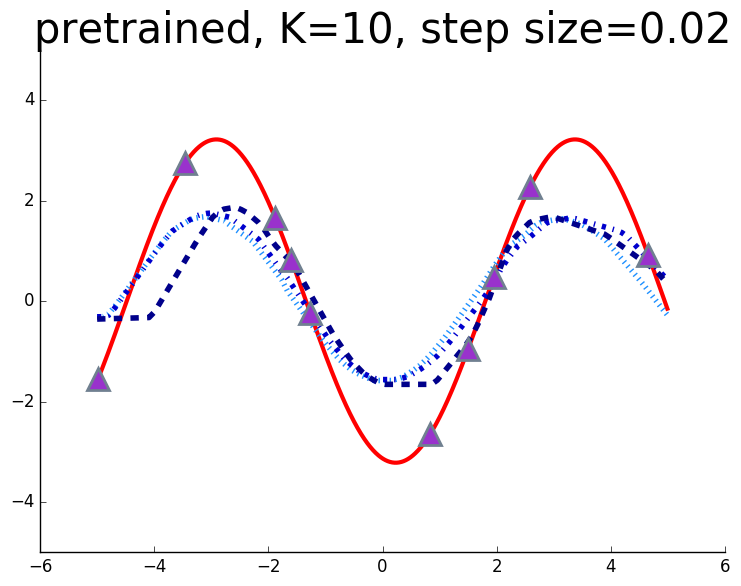}}
\put(3.0,4.0){\includegraphics[width=0.5\columnwidth]{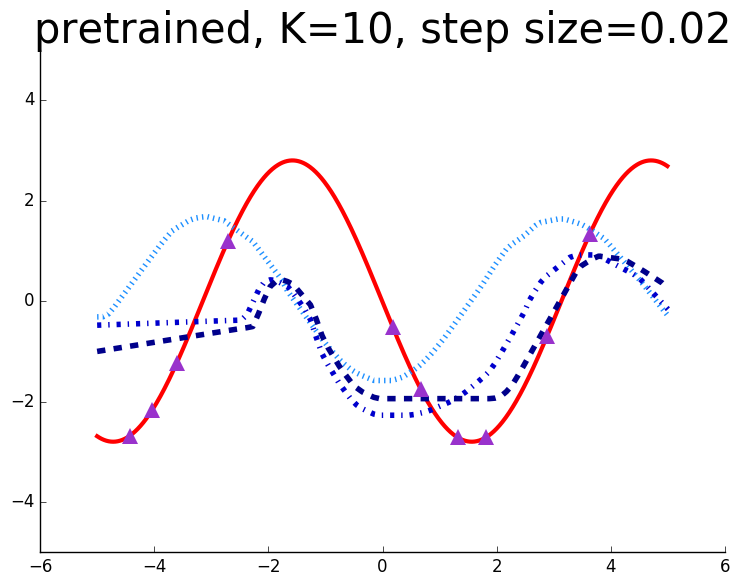}}

\put(0.0,-0.1){\includegraphics[width=2\columnwidth]{legend_sinusoid.pdf}}
\end{picture}
\caption{A random sample of qualitative results from the sinusoid regression task.
\label{fig:mamlqualapp}
\vspace{-0.5cm}
}
\end{figure*}

In Figure~\ref{fig:sinequant}, we show the full quantitative results of the MAML model trained on $10$-shot learning and evaluated on $5$-shot, $10$-shot, and $20$-shot. In Figure~\ref{fig:mamlqualapp}, we show the qualitative performance of MAML and the pretrained baseline on randomly sampled sinusoids.

\section{Additional Comparisons}
\label{app:comparisons}
In this section, we include more thorough evaluations of our approach, including additional multi-task baselines and a comparison representative of the approach of~\citet{rei}.

\subsection{Multi-task baselines}
The pretraining baseline in the main text trained a single network on all tasks, which we referred to as ``pretraining on all tasks''. To evaluate the model, as with MAML, we fine-tuned this model on each test task using $K$ examples. In the domains that we study, different tasks involve different output values for the same input. As a result, by pre-training on all tasks, the model would learn to output the average output for a particular input value. In some instances, this model may  learn very little about the actual domain, and instead learn about the range of the output space.

We experimented with a multi-task method to provide a point of comparison, where instead of averaging in the output space, we averaged in the parameter space.  To achieve averaging in parameter space, we sequentially trained $500$ separate models on $500$ tasks drawn from $p(\task)$. Each model was initialized randomly and trained on a large amount of data from its assigned task. We then took the average parameter vector across models and fine-tuned on 5 datapoints with a tuned step size. All of our experiments for this method were on the sinusoid task because of computational requirements. The error of the individual regressors was low: less than 0.02 on their respective sine waves. 

We tried three variants of this set-up. During training of the individual regressors, we tried using one of the following: no regularization, standard $\ell_2$ weight decay, and $\ell_2$ weight regularization to the mean parameter vector thus far of the trained regressors. The latter two variants encourage the individual models to find parsimonious solutions. When using regularization, we set the magnitude of the regularization to be as high as possible without significantly deterring performance. In our results, we refer to this approach as ``multi-task''. As seen in the results in Table~\ref{tbl:multitask_baselines}, we find averaging in the parameter space (multi-task) performed worse than averaging in the output space (pretraining on all tasks). This suggests that it is difficult to find parsimonious solutions to multiple tasks when training on tasks separately, and that MAML is learning a solution that is more sophisticated than the mean optimal parameter vector.

\begin{table}[t]
\caption{Additional multi-task baselines on the sinusoid regression domain, showing 5-shot mean squared error. The results suggest that MAML is learning a solution more sophisticated than the mean optimal parameter vector.}
\label{tbl:multitask_baselines}
\begin{center}
\begin{tabular}{c|c|c|c}
\hline
num. grad steps &  1   & 5 & 10\\
\hline
multi-task, no reg &  $4.19$ & $3.85$ & $3.69$ \\
\hline
multi-task, l2 reg & $7.18$ & $5.69$ & $5.60$ \\
\hline
multi-task, reg to mean $\theta$ & $2.91$ &$ 2.72$ & $2.71$ \\ % 0.001
\hline
pretrain on all tasks & $2.41$ & $2.23$ & $2.19$ \\
\hline
MAML (ours) & $\mathbf{0.67}$ & $\mathbf{0.38}$ & $\mathbf{0.35}$ \\
\hline
\end{tabular}
% this is to deal with awkward spacing...
%\vspace{-0.6cm}
\end{center}
\end{table}

\subsection{Context vector adaptation}
\citet{rei} developed a method which learns a context vector that can be adapted online, with an application to recurrent language models. The parameters in this context vector are learned and adapted in the same way as the parameters in the MAML model. To provide a comparison to using such a context vector for meta-learning problems,  we concatenated a set of free parameters $\mathbf{z}$ to the input $\inp$, and only allowed the gradient steps to modify $\mathbf{z}$, rather than modifying the model parameters $\theta$, as in MAML. For image inputs, $\mathbf{z}$ was concatenated channel-wise with the input image. We ran this method on Omniglot and two RL domains following the same experimental protocol. We report the results in Tables~\ref{tbl:rei_omniglot},~\ref{tbl:rei_pointmass}, and~\ref{tbl:rei_cheetah}. Learning an adaptable context vector performed well on the toy pointmass problem, but sub-par on more difficult problems, likely due to a less flexible meta-optimization.

\begin{table}
\begin{center}
\caption{5-way Omniglot Classification}
\label{tbl:rei_omniglot}
\begin{tabular}{c|c|c}
\hline
 & 1-shot & 5-shot\\
\hline
context vector & $94.9 \pm 0.9\%$ & $97.7 \pm 0.3\%$ \\
\hline
MAML & $\mathbf{98.7 \pm 0.4\%}$ &  $\mathbf{99.9 \pm 0.1\%}$ \\
\hline
\end{tabular}
\end{center}
\end{table}

\begin{table}
\caption{2D Pointmass, average return}
\label{tbl:rei_pointmass}
\begin{center}
\begin{tabular}{c|c|c|c|c}
\hline
num. grad steps & 0 & 1   & 2 & 3\\
\hline
context vector & $-42.42$ & $-13.90$ & $-5.17$ & $\mathbf{-3.18}$ \\
\hline
MAML (ours) & $-40.41$ & $\mathbf{-11.68}$ & $\mathbf{-3.33}$ & $-3.23$ \\
\hline
\end{tabular}
\end{center}
\end{table}

\begin{table}
\caption{Half-cheetah forward/backward, average return }
\label{tbl:rei_cheetah}
\begin{center}
\begin{tabular}{c|c|c|c|c}
\hline
num. grad steps & 0 & 1   & 2 & 3\\
\hline
context vector  & $-40.49$ & $-44.08$ & $-38.27$ & $-42.50$ \\
\hline
MAML (ours) & $-50.69$ & $\mathbf{293.19}$ & $\mathbf{313.48}$ & $\mathbf{315.65}$ \\
\hline
\end{tabular}
\end{center}
\end{table}

\end{document}